\newcommand{\PaperAuthors}{%
  {\large Kia Karbasi\textsuperscript{1}, Mahdi Saiedi\textsuperscript{2}, Kevin Hong\textsuperscript{1}, Gregory Pottie\textsuperscript{1}}\\[0.6em]
  {\normalsize \textsuperscript{1}University of California, Los Angeles}\\
  {\normalsize \textsuperscript{2}Sharif University of Technology}\\[0.3em]
  {\small \href{mailto:kiakarbasi@g.ucla.edu}{\texttt{kiakarbasi@g.ucla.edu}},\quad
  \href{mailto:msaiedi.042@gmail.com}{\texttt{msaiedi.042@gmail.com}}}\\
  {\small \href{mailto:kevinhong1167@g.ucla.edu}{\texttt{kevinhong1167@g.ucla.edu}},\quad
  \href{mailto:pottie@ee.ucla.edu}{\texttt{pottie@ee.ucla.edu}}}%
}
\newcommand{\PaperAuthorMetadata}{Kia Karbasi, Mahdi Saiedi, Kevin Hong, Gregory Pottie}
\documentclass[notitlepage,12pt]{jedm}

\usepackage[T1]{fontenc}
\usepackage[utf8]{inputenc}
\usepackage[english]{babel}
\usepackage{amsmath}
\usepackage{amssymb}
\usepackage{mathtools}
\usepackage{array}
\usepackage{booktabs}
\usepackage{tabularx}
\usepackage[table]{xcolor}
\usepackage{graphicx}
\usepackage{caption}
\usepackage{subcaption}
\usepackage{tikz}
\usetikzlibrary{arrows.meta,positioning}
\usepackage[section]{placeins}
\usepackage{url}
\usepackage{hyperref}

\providecommand{\PaperAuthors}{{\large Anonymous Author(s)}}
\providecommand{\PaperAuthorMetadata}{}

\hypersetup{
  colorlinks = true,
  urlcolor   = blue,
  linkcolor  = blue,
  citecolor  = blue,
  pdfauthor  = {\PaperAuthorMetadata},
  pdftitle   = {Does Deep Knowledge Tracing Learn Semantic Relations Between Knowledge Components?}
}
\providecommand{\chapter}[1]{}
\providecommand{\DataTableHeaderRows}[1]{}
\makeatletter
\@ifundefined{KV@Gin@alt}{\define@key{Gin}{alt}{}}{}
\makeatother

\begin{document}

\title{Does Deep Knowledge Tracing Learn\\Semantic Relations Between\\Knowledge Components?}
\date{}
\author{\PaperAuthors}
\maketitle

\begin{abstract}
Deep Knowledge Tracing (DKT) is a knowledge tracing (KT) model that updates a shared latent state from sequences of interactions, each pairing a knowledge component (KC) with a response. An interaction with one KC can therefore affect predictions for other KCs. We ask: Are the KC relations learned by DKT semantic? We construct text-derived semantic KC representations using nine pipelines to test whether the conclusions hold across alternative ways of constructing those representations.

The first test compares DKT-derived KC relation graphs with semantic relation structures induced by pairwise cosine similarities among the KC representations. Across the evaluated datasets, embedding models, and pipelines, the two structures align only weakly and inconsistently. The second test introduces semantic information into DKT; the pretrained text-derived representations do not consistently outperform the standard DKT baseline.

In the reduced-data experiments, semantic initialization yields a higher test area under the ROC curve (AUC) than the baseline at eight of nine nested sample sizes for ASSISTments 2017. KC-supervised fine-tuned BERT produces small positive AUC differences in some matched comparisons, but the study does not establish their stability across runs. These effects remain limited and inconsistent. The findings suggest that behavioral, temporal, structural, or psychometric signals may also shape DKT's learned relations. DKT-derived relations are not primarily semantic: the tested text-derived semantic structures do not strongly recover the DKT-derived relation structure.\\

{\parindent0pt
\textbf{Keywords:} knowledge tracing, semantic embeddings, question text, Deep Knowledge Tracing
}
\end{abstract}

\section{Introduction}

Knowledge tracing (KT) models estimate the probability that a student will answer the next question correctly from the student's prior interaction history~\cite{corbett_knowledge_1995,piech_deep_2015}. In datasets such as ASSISTments, each interaction is linked to a question and one or more knowledge components (KCs), and the model must infer latent mastery over time. Bayesian Knowledge Tracing (BKT) does this with an explicit hidden-state probabilistic model, whereas Deep Knowledge Tracing (DKT) updates a shared latent state from sequences of KC-response interactions with recurrent networks. This shared-state design lets us ask which relations among questions and KCs support prediction.

In the original DKT example, responses to \(x\)-intercept exercises can update the model's predicted probability of a correct response to \(y\)-intercept exercises~\cite{piech_deep_2015}. Because these KCs cover closely related mathematical content, this cross-KC behavior motivates our central question: \emph{Are the KC relations learned by DKT semantic?}

In this paper, we evaluate the hypothesis that question text carries information about what a student is being asked to do and that DKT-derived KC relations might therefore be semantic. We test that hypothesis under controlled conditions rather than assume that text-derived semantic structure aligns with DKT-derived relations or that semantic initialization improves DKT's predictive performance.

Educational questions are not arbitrary identifiers. They contain instructions, mathematical language, and cues about the skill being assessed. Modern text representation methods are designed to place semantically similar texts near one another in vector space~\cite{devlin_bert_2019,reimers_sentence-bert_2019,gao_simcse_2021}. One hypothesis is that these semantic representations could provide a better inductive bias than sparse ID-based inputs.

In this study, we use nine pipelines to map question text into pretrained KC-response representations. The alternatives test whether the findings hold across different ways of constructing semantic KC structure.

We assess the semantic-relation question through two complementary tests. The first and most direct test compares the DKT-derived KC relation graph with the semantic KC relation structure induced by pairwise cosine similarities among text-derived KC representations. The second test replaces the default input embedding table in pyKT's DKT implementation with pretrained embeddings derived from question text. The goal is to determine whether semantic information in question text, when mapped into KC-response representations, improves DKT's predictive performance.

The principal result is weak structural alignment. DKT-derived KC relations are only weakly aligned with the text-derived semantic structures evaluated in this study. Across the pipelines studied here, pretrained text-derived representations also do not consistently outperform the standard DKT baseline. This result constrains what semantic similarity can explain under this setup and suggests that behavioral, temporal, structural, or psychometric signals may also shape DKT's learned relations. That interpretation is consistent with later KT work that couples question information to additional structural or contextual signals rather than relying on text alone~\cite{ghosh_context-aware_2020,pardos_imputing_2017,wang_seep_2022}.

We investigate the following research questions:
\begin{itemize}
    \item \textbf{RQ1.} Are the KC relations learned by DKT semantic?
    \item \textbf{RQ2.} Does introducing semantic information improve DKT's modeling performance?
    \begin{itemize}
        \item \textbf{RQ2.1.} Does semantic initialization improve DKT's predictive performance in the standard full-data setting?
        \item \textbf{RQ2.2.} Does semantic initialization improve DKT's modeling performance when the amount of behavioral data is reduced?
    \end{itemize}
\end{itemize}
The graph-alignment analysis directly answers \textbf{RQ1}. The full-data and reduced-data comparisons answer \textbf{RQ2.1} and \textbf{RQ2.2}, respectively. The fine-tuned BERT experiments provide a mechanism analysis under \textbf{RQ2}.

In this paper, we make four contributions. First, we formulate a direct semantic hypothesis for DKT and test whether semantic KC representations recover the relation structure that DKT appears to use. Second, we develop and compare nine pipelines for constructing KC-response embeddings from question text. Third, we evaluate those embeddings both under the standard training setup and under reduced-data regimes. Fourth, under RQ2, we analyze whether aligning the text encoder with KC labels improves DKT's predictive performance.

The rest of this paper is structured as follows. We first review the background most relevant to the study, then describe the methodology, report the results, and discuss what those results imply for KT.

\section{Background and Related Work}

\subsection{Knowledge Tracing and DKT}

DKT updates a shared latent state from sequences of KC-response interactions and therefore implicitly models relationships among KCs. Evidence that a student knows or does not know one KC can affect the model's prediction concerning another KC.

The original DKT work also introduced an influence-score method for relating KCs based on how a correct response to one KC changes the predicted probability for another~\cite{piech_deep_2015}. Later work examined this graph-extraction procedure and proposed a modified repeated-response method for extracting exercise-relation graphs~\cite{hong2025extracting}.

Later KT research suggests that the KC relations useful for prediction need not be semantic. DKT's gains were followed by critiques showing that model behavior can be unstable or hard to interpret, and that preprocessing and benchmark details can materially affect conclusions~\cite{khajah_deep_2016,xiong_going_2016,yeung_addressing_2018}. More recent architectures often improve predictive performance by modeling memory, recency, attention, question difficulty, or question-level structure more explicitly~\cite{zhang_dynamic_2017,pandey_self-attentive_2019,sonkar_qdkt_2020,ghosh_context-aware_2020}. This literature motivates two tests: whether semantic structure corresponds to DKT-derived KC relations and whether semantic initialization improves DKT's predictive performance without a more elaborate architecture.

Relation-aware KT models address a neighboring prediction problem by supplying, learning, or propagating over relations among concepts and exercises. Prerequisite-driven and structure-based models encode directed or multiple concept relations, graph neural KT treats knowledge structure as a graph, and heterogeneous or hierarchical graph models represent question--skill and exercise relations~\cite{chenPrerequisiteDrivenDeepKnowledge2018,nakagawaGraphbasedKnowledgeTracing2019,tongStructureBasedKnowledgeTracing2020,luo_heterogeneous_2022,tong_hgkt_2022,duanMoreAccurateInterpretable2024}. Those models use relation structure to improve predictive performance or interpret model behavior. This study instead asks a diagnostic question: whether the relation matrix extracted from baseline DKT by the original influence-score procedure aligns with text-derived KC cosine geometry.

\subsection{Pretrained Representations of Educational Questions}

Prior work has augmented KT models with exercise-side information, built exercise-aware architectures, proposed specialized question representations, and pretrained question embeddings for educational prediction~\cite{su_exercise-enhanced_2018,liu_ekt_2019,yin_quesnet_2019,wang_seep_2022}. At the same time, KC-imputation work suggests that question text may be insufficient by itself and that surrounding interaction context can be equally important~\cite{pardos_imputing_2017}. Together, these findings motivate a controlled test of whether question text improves DKT's predictive performance.

\subsection{Pretrained Embeddings as KT Inputs}

Distributional and contextual embedding methods map text into dense vector spaces in which lexical, topical, and semantic relations are partially preserved~\cite{mikolov_distributed_2013,bojanowski_enriching_2017,pennington_glove_2014,devlin_bert_2019,reimers_sentence-bert_2019,gao_simcse_2021}. Sentence-level embedding methods are especially relevant here because the study compares full questions and groups of questions rather than isolated tokens. The representation flow is:
\[
\text{question text} \rightarrow \text{question embedding} \rightarrow \text{KC representation} \rightarrow \text{KT input}.
\]

If semantic structure were aligned with the dependencies useful for KT, then these dense representations could improve generalization, reduce sparsity, and increase predictive performance when less behavioral data are available. That expectation is consistent with related work. Exercise-enhanced and exercise-aware KT models incorporate richer representations of questions rather than treating them as anonymous IDs~\cite{su_exercise-enhanced_2018,liu_ekt_2019}. Question-aware KT architectures likewise treat question representation as a central design choice~\cite{sonkar_qdkt_2020,ghosh_context-aware_2020}.

Related educational question-representation work, such as QuesNet, argues that test questions are heterogeneous objects containing both linguistic content and domain logic, which may not be well served by generic off-the-shelf embeddings alone~\cite{yin_quesnet_2019}. More directly, pretraining question embeddings for KT with structural constraints has been shown to improve predictive performance in settings where question--concept relations and difficulty are modeled explicitly~\cite{wang_seep_2022}.

At the same time, the literature also suggests reasons this hypothesis might fail. Pardos and Dadu show that for KC imputation, contextual information from neighboring problems can be as predictive as the problem text itself~\cite{pardos_imputing_2017}. The semantics relevant to educational prediction may therefore be distributed across wording, curriculum structure, and usage context rather than residing only in the text. In ASSISTments-style data, operational KC labels, student behavior, and dataset construction may not align perfectly with pure linguistic similarity.

Accordingly, we use semantic representations as a controlled probe of DKT. We test whether they align with DKT-derived relations and whether semantic initialization improves DKT's predictive performance.

\section{Methodology}

\subsection{Data Sources and Text Coverage}

This study uses the ASSISTments 2009, 2012, and 2017 KT benchmarks~\cite{abdelrahmanKnowledgeTracingSurvey2023,fengAddressingAssessmentChallenge2009,priharExploringCommonTrends2022}. Because semantic KC representations require question text, the analyses also rely on the Problem Bodies resource, which provides text for only a subset of ASSISTments items. Only questions and KCs with matched text receive text-derived representations.

The text-linked subset is much smaller than the full interaction log, especially at the question level; this coverage constraint shapes every text-based analysis in the study. Table~\ref{tab:chapter5_pb_data} reports the number of unique questions and KCs in each interaction log together with the corresponding counts in the matched Problem Bodies subset.

\begin{table}[!htbp]
    \centering
    \small
    \setlength{\tabcolsep}{6pt}
    \renewcommand{\arraystretch}{1.1}
    \caption{Question and KC counts in the ASSISTments logs and in the matched Problem Bodies subset used for text-based analyses in this study.}
    \label{tab:chapter5_pb_data}
    \DataTableHeaderRows{1}
    \begin{tabular}{lcccc}
        \toprule
        Dataset & \# Questions & \# Questions in PB & \# KCs & \# KCs in PB \\
        \midrule
        ASSISTments 2009 & 26,688 & 7,836 & 123 & 76 \\
        ASSISTments 2012 & 179,999 & 32,792 & 265 & 162 \\
        ASSISTments 2017 & 3,162 & 638 & 102 & 68 \\
        \bottomrule
    \end{tabular}
\end{table}

The ASSISTments 2009 Problem Bodies count treats \texttt{311} and \texttt{311.0} as two matched KC identifier strings; numeric normalization merges them, so the DKT and RQ1 matrices use 75 modeled KCs. Using the modeled KC counts, the matched-question counts in Table~\ref{tab:chapter5_pb_data} correspond to approximately 104.5, 202.4, and 9.4 questions per KC for ASSISTments 2009, 2012, and 2017, respectively.

All text-based analyses operate on trimmed text-linked datasets. Questions and KCs without matched Problem Bodies text are excluded from the semantic-representation and DKT comparison pipeline. The DKT baselines, semantic-initialization models, reduced-data experiments, BERT downstream experiments, and graph-alignment analyses use the same dataset representation. Throughout this paper, \emph{full-data} refers to the complete trimmed dataset rather than the original unfiltered interaction log.

When dataset text availability is compared below, \emph{largest text-linked question pool} and \emph{smallest text-linked question pool} refer to the absolute matched-question counts in Table~\ref{tab:chapter5_pb_data}, not to the proportion of all questions or KCs that are matched.

\subsection{Constructing Semantic KC Representations}

\begin{figure}[!htbp]
    \centering
    \begin{tikzpicture}[
        node distance=1.25cm,
        box/.style={draw=black!65, rounded corners=2pt, fill=blue!5, minimum height=1.25cm, align=center},
        flow/.style={-{Latex[length=2.5mm]}, line width=0.8pt, draw=blue!60!black}
    ]
        \node[box, text width=4.0cm] (question) {\textbf{Question text}\\``What is the \(y\)-intercept of \(4x+2y=12\)?''};
        \node[box, text width=2.4cm, right=of question] (encoder) {\textbf{Embedding model}\\ \(f_{\theta}\)};
        \node[box, text width=2.7cm, right=of encoder] (vector) {\textbf{Question vector}\\ \(e_q\in\mathbb{R}^{d}\)};
        \draw[flow] (question) -- (encoder);
        \draw[flow] (encoder) -- (vector);
    \end{tikzpicture}
    \caption{Question text is mapped by an embedding model to a dense question vector.}
    \label{fig:text_informed_question_embeddings}
    \Description{Three boxes connected from left to right contain the example question ``What is the \(y\)-intercept of \(4x+2y=12\)?'', the embedding model \(f_{\theta}\), and the output question vector \(e_q\in\mathbb{R}^{d}\).}
\end{figure}
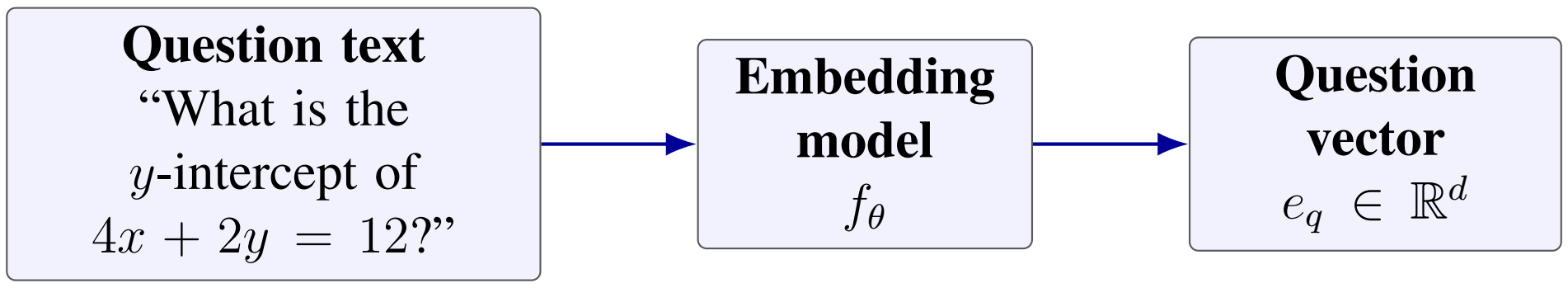

Each pipeline maps the question texts associated with a KC to two response-specific interaction vectors. The construction has three steps. The first is the \emph{question representation step}, in which raw question text is mapped into a dense vector representation. The second is the \emph{KC construction step}, in which one or more question-level representations are transformed into a single KC-level vector. The third is the \emph{response-state construction step}, in which the KC representation is converted into the two response-specific interaction embeddings required by DKT. The initial question-to-embedding step is illustrated in Figure~\ref{fig:text_informed_question_embeddings}.

We evaluate sentence-transformer models, commercial embedding APIs, and dataset-specific fine-tuned BERT encoders through nine semantic-representation pipelines. Each pipeline produces the two response-specific vectors required by DKT but combines the three construction steps differently.

\subsubsection{Embedding Models Used in the Semantic Representation Pipelines}

\begin{table}[!htbp]
    \centering
    \small
    \setlength{\tabcolsep}{4pt}
    \renewcommand{\arraystretch}{1.1}
    \caption{Embedding models used in the semantic representation pipelines. Dimensionalities are those used in the evaluated configurations.}
    \label{tab:chapter5_embedding_models}
    \DataTableHeaderRows{1}
    \begin{tabularx}{\textwidth}{>{\raggedright\arraybackslash}p{0.18\textwidth}>{\raggedright\arraybackslash}X>{\centering\arraybackslash}p{0.16\textwidth}>{\raggedright\arraybackslash}p{0.18\textwidth}}
        \toprule
        Family & Model & Dimensionality & Dataset Scope \\
        \midrule
        OpenAI & \path{text-embedding-3-small} & 1,536 & All three \\
        OpenAI & \path{text-embedding-3-large} & 3,072 & All three \\
        OpenAI & \path{text-embedding-ada-002} & 1,536 & All three \\
        Sentence-\newline Transformer & \path{all-MiniLM-L6-v2} & 384 & All three \\
        Sentence-\newline Transformer & \path{all-mpnet-base-v2} & 768 & All three \\
        Sentence-\newline Transformer & \path{multi-qa-mpnet-base-dot-v1} & 768 & All three \\
        Sentence-\newline Transformer & \path{all-MiniLM-L12-v2} & 384 & All three \\
        Sentence-\newline Transformer & \path{paraphrase-MiniLM-L6-v2} & 384 & All three \\
        Cohere & \path{embed-v4.0} & 1,536 & All three \\
        Cohere & \path{embed-english-v3.0} & 1,024 & All three \\
        Cohere & \path{embed-multilingual-v3.0} & 384 & All three \\
        Fine-tuned BERT & ASSISTments 2009 KC classifier & 768 & 2009 only \\
        Fine-tuned BERT & ASSISTments 2012 KC classifier & 768 & 2012 only \\
        \bottomrule
    \end{tabularx}
\end{table}

The pipeline design supports the same nine constructions across embedding backbones while accommodating backbone-specific dimensionality and context-window limits.

\subsection{Nine Representation Pipelines}

The nine pipelines form a conceptual \(3\times3\) design. The KC-construction dimension treats a KC as a semantic prototype or centroid (\emph{average}), a concrete exemplar (\emph{sample}), or a holistic document (\emph{stuff window}). The response-state dimension treats response state as polarity along a representational axis (\emph{mirror}), a separately appended attribute (\emph{stack antonym}), or a condition inserted into the text before embedding (\emph{add words}). Figure~\ref{fig:chapter5_pipeline_design} maps the nine pipelines to these two dimensions.

\begin{figure}[!htbp]
    \centering
    \resizebox{\linewidth}{!}{%
    \begin{tikzpicture}[
        cell/.style={draw=black!55, minimum width=3.3cm, minimum height=1.25cm, align=center, fill=blue!4},
        rowhead/.style={draw=black!55, minimum width=3.7cm, minimum height=1.25cm, align=center, fill=black!6},
        colhead/.style={draw=black!55, minimum width=3.3cm, minimum height=1.1cm, align=center, fill=blue!12},
        corner/.style={draw=black!55, minimum width=3.7cm, minimum height=1.1cm, align=center, fill=black!10}
    ]
        \node[corner] at (0,0) {\textbf{KC construction}\\\textbf{\(\times\) response state}};
        \node[colhead] at (3.5,0) {\textbf{Mirror}\\polarity};
        \node[colhead] at (7.0,0) {\textbf{Stack antonym}\\appended attribute};
        \node[colhead] at (10.5,0) {\textbf{Add words}\\conditioned text};

        \node[rowhead] at (0,-1.3) {\textbf{Prototype}\\average};
        \node[cell] at (3.5,-1.3) {\textbf{P1}};
        \node[cell] at (7.0,-1.3) {\textbf{P2}};
        \node[cell] at (10.5,-1.3) {\textbf{P7}};

        \node[rowhead] at (0,-2.6) {\textbf{Exemplar}\\sample};
        \node[cell] at (3.5,-2.6) {\textbf{P3}};
        \node[cell] at (7.0,-2.6) {\textbf{P4}};
        \node[cell] at (10.5,-2.6) {\textbf{P8}};

        \node[rowhead] at (0,-3.9) {\textbf{Holistic document}\\stuff window};
        \node[cell] at (3.5,-3.9) {\textbf{P5}};
        \node[cell] at (7.0,-3.9) {\textbf{P6}};
        \node[cell] at (10.5,-3.9) {\textbf{P9}};
    \end{tikzpicture}%
    }
    \caption{Conceptual \(3\times3\) design of the nine semantic-representation pipelines.}
    \label{fig:chapter5_pipeline_design}
    \Description{Three-by-three matrix mapping prototype, exemplar, and holistic-document KC constructions to mirror, stack-antonym, and add-words response-state constructions.}
\end{figure}

The same pipelines can also be grouped by their execution order. In this notation, student action (SA) denotes the response state attached to a KC:
\begin{enumerate}
    \item \textbf{Question $\rightarrow$ KC $\rightarrow$ SA pipelines (Pipelines 1--4):} raw question text is embedded first, KC embeddings are then created either by averaging all questions in a KC or by sampling a single representative question, and finally correctness is added using either a mirror transformation or a concatenated antonym embedding.
    \item \textbf{KC $\rightarrow$ Question $\rightarrow$ SA pipelines (Pipelines 5--6):} each KC is represented directly by concatenating as many of its question texts as will fit in a model-dependent context window and embedding that single stuffed text. Correctness is then added using either the mirror or stack-antonym transformation.
    \item \textbf{SA $\rightarrow$ Question $\rightarrow$ KC pipelines (Pipelines 7--8):} correctness is injected before embedding by prefixing each question text with repeated \texttt{CORRECT} or \texttt{INCORRECT} tokens. The modified questions are then embedded and aggregated into separate correct and incorrect KC vectors using either averaging or single-question sampling.
    \item \textbf{SA $\rightarrow$ KC $\rightarrow$ Question pipeline (Pipeline 9):} response state is inserted into each question text, the modified texts are concatenated into separate response-specific KC windows, and each window is embedded once.
\end{enumerate}

\begin{table}[!htbp]
    \centering
    \footnotesize
    \setlength{\tabcolsep}{2.5pt}
    \renewcommand{\arraystretch}{1.1}
    \caption{Nine evaluated pipelines for generating response-specific KC embeddings. SA denotes student action, the response state attached to a KC.}
    \label{tab:text_informed_pipelines}
    \DataTableHeaderRows{1}
    \begin{tabularx}{\textwidth}{c >{\raggedright\arraybackslash}p{2.7cm} >{\raggedright\arraybackslash}p{1.8cm} >{\raggedright\arraybackslash}X >{\raggedright\arraybackslash}p{1.8cm} >{\raggedright\arraybackslash}p{2.2cm}}
        \toprule
        Pipeline & Execution order & Question-text operation & Question representation step & KC construction step & Response-state construction step \\
        \midrule
        P1 & question $\rightarrow$ KC $\rightarrow$ SA & raw question text & embed raw question & average & mirror \\
        P2 & question $\rightarrow$ KC $\rightarrow$ SA & raw question text & embed raw question & average & stack antonym \\
        P3 & question $\rightarrow$ KC $\rightarrow$ SA & raw question text & embed raw question & sample single question & mirror \\
        P4 & question $\rightarrow$ KC $\rightarrow$ SA & raw question text & embed raw question & sample single question & stack antonym \\
        P5 & KC $\rightarrow$ question $\rightarrow$ SA & raw question text & embed stuffed KC text & stuff window & mirror \\
        P6 & KC $\rightarrow$ question $\rightarrow$ SA & raw question text & embed stuffed KC text & stuff window & stack antonym \\
        P7 & SA $\rightarrow$ question $\rightarrow$ KC & add words & embed modified question & average & encoded in text \\
        P8 & SA $\rightarrow$ question $\rightarrow$ KC & add words & embed modified question & sample single question & encoded in text \\
        P9 & SA $\rightarrow$ KC $\rightarrow$ question & add words & embed stuffed modified KC text & stuff window & encoded in text \\
        \bottomrule
    \end{tabularx}
\end{table}

\FloatBarrier

Two construction details matter for interpreting these pipelines. First, the \texttt{mirror} strategy keeps one response-specific KC vector as the base embedding \(e\) and maps the other to \(-e\). Second, the \texttt{stack\_antonym} strategy concatenates the KC vector with a separately embedded response marker, doubling the dimensionality of the final response-specific input. By contrast, the \texttt{add\_words} pipelines do not apply a later response-state transformation at the KC level; response state is already embedded in the modified question text itself. In the experiments reported here, those prefixes are repeated twice before embedding.

Stuff-window construction first chooses the model-dependent context window, then concatenates as many question texts as fit, and embeds the resulting document once. OpenAI models use a larger window than the sentence-transformer models. Table~\ref{tab:text_informed_pipelines} summarizes the nine combinations.

Averaging pools the question embeddings associated with a KC by arithmetic mean. Sampling selects one representative question and uses its embedding as the KC representation. Stuffing concatenates as many associated question texts as fit in the model-dependent window and embeds the document once.

Mirroring maps the two response states to opposite directions along the same representational axis. Stack-antonym concatenates a KC embedding with a separately embedded response marker. Add-words injects response state into each question text before embedding. Together, these families test whether the conclusions depend on where response state enters representation construction.

\FloatBarrier

\subsection{DKT-Derived KC Relation Graphs}

The relation graphs were produced for this study by applying the original DKT influence-score method to the same baseline DKT models whose predictive results are reported in Table~\ref{tab:chapter5_results_combined}. Following Piech et al.~\cite{piech_deep_2015}, we assign an influence score \(J_{ij}\) to each directed source--target pair:
\[
J_{ij} = \frac{y^{(j \mid i)}}{\sum_k y^{(j \mid k)}},
\]
where \(y^{(j \mid i)}\) is the correctness probability assigned by the trained DKT model to target KC \(j\) on the second timestep, given a correct response to source KC \(i\) on the first timestep. The denominator sums over every source KC \(k\) in the same trimmed text-linked KC set. We use the resulting complete weighted relation matrix over that set as the DKT-derived comparison target for RQ1.

\subsection{Semantic Relation Structure and Alignment Metrics}

We assess the semantic-relation question by comparing two structures over the same ordered set of text-linked KCs: the DKT-derived KC relation matrix and the geometry induced by the pretrained KC embeddings. For the RQ1 semantic graph, we use the first \(N\) rows of each \(2N\)-row pretrained interaction tensor to obtain one vector per KC. Pairwise cosine similarity over these KC vectors produces a symmetric semantic relation matrix.

Each dataset-specific KC index (\texttt{keyid2idx.json}) orders the trimmed text-linked KCs. The DKT-derived relation matrix and semantic cosine-similarity matrix are compared over exactly that same ordered set. KCs outside the trimmed text-linked set are excluded from the comparison.

The extracted DKT relation matrix is directed, whereas cosine similarity is symmetric. Reciprocal DKT weights are averaged to form a symmetric weighted matrix, and the diagonal is set to zero.

Pearson and Spearman correlations compare corresponding strict-upper-triangle entries of the symmetrized DKT matrix and the semantic cosine-similarity matrix; Spearman first replaces the values by their ranks. For top-\(k\) edge overlap, each matrix contributes its \(k\) highest-weight strict-upper-triangle edges at 1\%, 5\%, 10\%, and 20\% of the available edges, and overlap is normalized as \(\lvert A \cap B \rvert/k\). Exactly \(k\) entries are selected without expanding boundary ties.

As a non-text reference, we also construct a symmetric KC co-exposure matrix derived from the same training sequences used to fit DKT, counting each unordered KC pair at most once per student sequence and setting the diagonal to zero. The comparator collapses response order and correctness into student-level co-exposure and is compared with the DKT-derived matrix over the same KC set, ordering, strict-upper-triangle entries, and top-\(k\) conventions used for the semantic comparisons.

We use \emph{weak alignment} descriptively: the reported correlations and edge overlaps are small in absolute value. We do not use inferential significance testing to make claims about graph recovery, and we do not estimate how stable the extracted DKT relation matrices are across retraining runs. The alignment magnitudes should therefore be read as descriptive comparisons rather than calibrated recoverability estimates.

\subsection{Introducing Semantic Information into DKT}

The baseline used in this study is the DKT model provided by the pyKT library~\cite{liuPyKTPythonLibrary2023}. Each student interaction is represented as a discrete KC-response state and indexes an embedding lookup table. Let \(i_t\) denote the interaction index assigned to the KC-response pair \((c_t,r_t)\). For \(N\) KCs, the embedding table contains one row for each of the \(2N\) possible KC-response states.

In the standard pyKT DKT setup, this interaction embedding matrix is randomly initialized and learned during training. If \(E \in \mathbb{R}^{2N \times d_{\mathrm{base}}}\) denotes the embedding table and \(e_{i_t}\) its \(i_t\)-th row, then the recurrent model receives
\[
x_t = e_{i_t}, \qquad E \sim \text{random initialization},
\]
where \(d_{\mathrm{base}}\) is the baseline embedding dimension. The pyKT DKT configuration used in this study defaults to \(d_{\mathrm{base}}=200\) and zero L2 weight decay. The baseline comparison is therefore between a randomly initialized learned interaction-embedding table and a pretrained interaction-embedding table with the same role in the model.

Our modification replaces that randomly initialized lookup table with pretrained interaction embeddings constructed by the evaluated pipeline variants. Let \(\tilde{E} \in \mathbb{R}^{2N \times d_{\mathrm{sem}}}\) denote the pretrained matrix whose rows are the response-specific KC embeddings generated from question text. The modified model uses
\[
x_t = \tilde{e}_{i_t}, \qquad \tilde{E} =
\begin{bmatrix}
\tilde{e}_{1} \\
\vdots \\
\tilde{e}_{2N}
\end{bmatrix},
\]
where each row is supplied by one of the nine semantic-representation pipelines. The pretrained matrix sets the lookup-table dimensionality and initializes its weights.

\begin{figure}[!htbp]
    \centering
    \begin{subfigure}{0.48\linewidth}
        \centering
        \includegraphics[width=\linewidth]{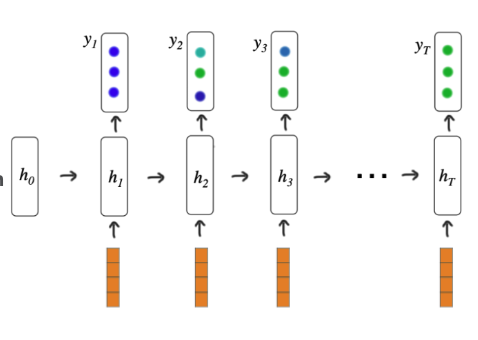}
        \caption{Pretrained embedding initialization}
    \end{subfigure}\hfill
    \begin{subfigure}{0.48\linewidth}
        \centering
        \includegraphics[width=\linewidth]{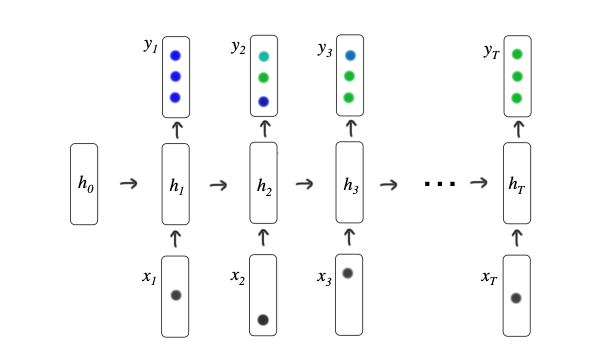}
        \caption{Baseline DKT with randomly initialized interaction embeddings}
    \end{subfigure}
    \caption{Pretrained embedding initialization. The panels compare the text-derived KC-response initialization with the baseline DKT interaction representation based on randomly initialized embeddings.}
    \label{fig:chapter5_pretrained_embedding_initialization}
    \Description{Both panels show a recurrent DKT sequence from hidden state \(h_0\) through \(h_T\), with an output vector above each state. The left panel supplies a pretrained text-derived KC-response embedding at each time step, whereas the right panel uses the baseline interaction representation with randomly initialized embeddings.}
\end{figure}

Each pipeline therefore supplies two response-specific vectors per KC to populate the \(2N\)-row table. Figure~\ref{fig:chapter5_pretrained_embedding_initialization} depicts this shared replacement.

We evaluate predictive performance with area under the ROC curve (AUC) and accuracy.

\subsection{Architecture and Robustness Variants}

We evaluate several architecture variants as robustness analyses for RQ2.1. Each retains pretrained embedding initialization while changing a different part of DKT.

\paragraph{Projection Bottleneck}

\begin{figure}[!htbp]
    \centering
    \includegraphics[width=0.72\linewidth]{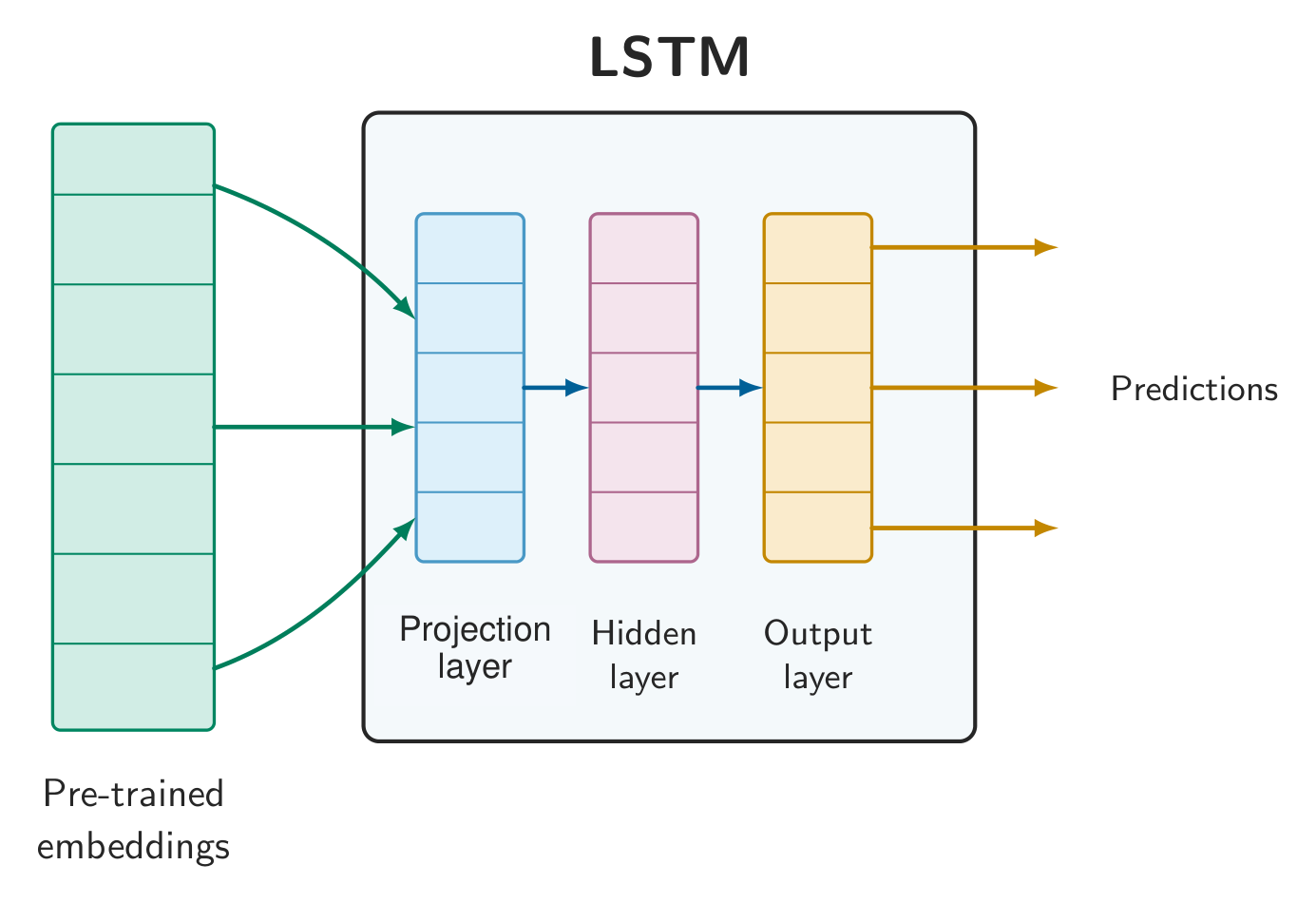}
    \caption{Projection bottleneck applied to the pretrained-input DKT model.}
    \Description{Diagram of an LSTM-based DKT model with pretrained embeddings passed through a learned projection layer before the recurrent hidden and output layers.}
    \label{fig:chapter5_projection_bottleneck}
\end{figure}

The projection-bottleneck variant inserts a learned projection layer before the recurrent model. As shown in Figure~\ref{fig:chapter5_projection_bottleneck}, this adds a learned compression stage between the pretrained interaction embedding and the recurrent computation. If \(x_t \in \mathbb{R}^{d_{\mathrm{sem}}}\) denotes the pretrained interaction embedding, the projected representation is
\[
z_t = W_{\mathrm{proj}} x_t + b_{\mathrm{proj}},
\]
where \(W_{\mathrm{proj}} \in \mathbb{R}^{d_{\mathrm{proj}} \times d_{\mathrm{sem}}}\). In the bottleneck setting, \(d_{\mathrm{proj}} < d_{\mathrm{sem}}\), so the projection compresses the pretrained representation before sequence modeling. This serves two purposes. First, it adapts high-dimensional pretrained embeddings to the dimensionality expected by the recurrent model. Second, it forces the model to compress the semantic representation into a smaller space tailored to the prediction task, which may help filter irrelevant semantic variation.

Table~\ref{tab:chapter5_embedding_models} reports each backbone's semantic dimension. Mirror and add-words pipelines retain that width for each response-specific input, whereas stack-antonym concatenates two backbone-width vectors and therefore doubles it.

\FloatBarrier

\paragraph{Regularization}

The regularization variants tune dropout and, where configured, L2 weight decay. The 2009 direct-initialization condition is a dropout-regularized configuration; L2 weight decay is not part of the default pyKT DKT setting. These conditions test whether stronger regularization improves stability or generalization with high-dimensional pretrained inputs.

\paragraph{Adaptive Recurrent Capacity}

\begin{figure}[!htbp]
    \centering
    \begin{tikzpicture}[
        node distance=1.2cm,
        box/.style={draw=black!65, rounded corners=2pt, fill=blue!5, minimum height=1.35cm, align=center},
        flow/.style={-{Latex[length=2.5mm]}, line width=0.8pt, draw=blue!60!black}
    ]
        \node[box, text width=2.3cm] (input) {\textbf{Input}\\ \(u_t\)};
        \node[box, text width=5.2cm, right=of input] (lstm) {\textbf{LSTM recurrent block}\\Configuration varies in hidden size, depth, and dropout};
        \node[box, text width=2.3cm, right=of lstm] (output) {\textbf{Output}\\ \(h_t\)};
        \draw[flow] (input) -- (lstm);
        \draw[flow] (lstm) -- (output);
    \end{tikzpicture}
    \caption{Recurrent-capacity configurations in the pretrained-input DKT model.}
    \Description{Diagram of an LSTM-based DKT model whose recurrent configurations vary in hidden size, depth, and dropout.}
    \label{fig:chapter5_adaptive_recurrent_capacity}
\end{figure}
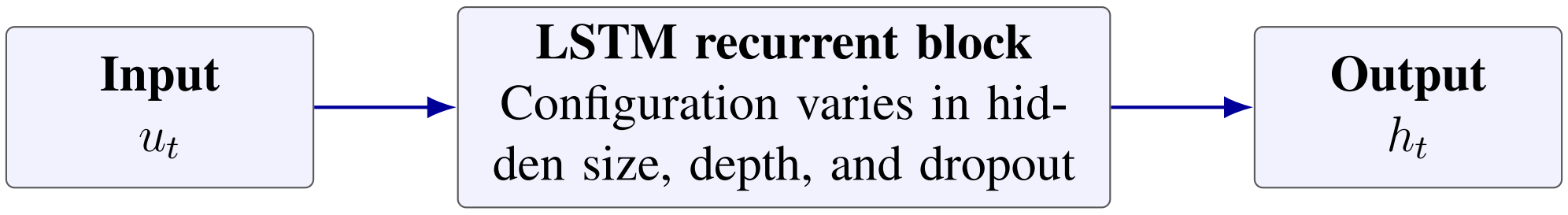

The adaptive-capacity variant, illustrated in Figure~\ref{fig:chapter5_adaptive_recurrent_capacity}, changes the recurrent block rather than semantic input construction. We vary hidden size, recurrent depth, and dropout across configurations. In abstract form, the recurrent update becomes
\[
h_t = \mathrm{LSTM}_{\theta}(u_t, h_{t-1}),
\]
where \(u_t = x_t\) in the direct-input case and \(u_t = z_t\) in the projected-input case, while \(\theta\) is selected from recurrent configurations with different hidden widths, depths, and dropout settings.

\paragraph{Semantic Output Layer}

\begin{figure}[!htbp]
    \centering
    \includegraphics[width=0.72\linewidth]{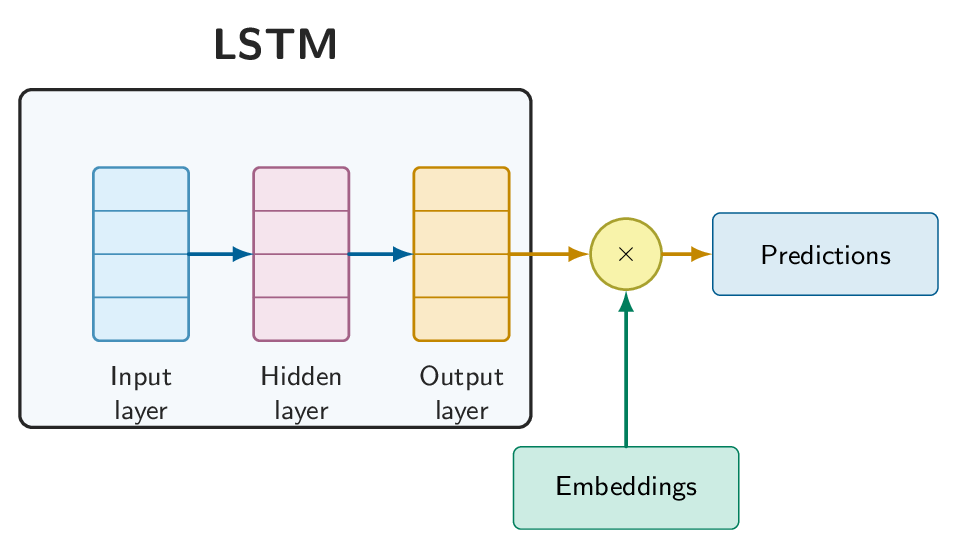}
    \caption{Semantic output layer in the pretrained-input DKT model.}
    \Description{Diagram of an LSTM-based DKT model whose prediction head combines the recurrent output with semantic embeddings before producing predictions.}
    \label{fig:chapter5_semantic_output_layer}
\end{figure}

The final architecture variant is the \emph{Semantic Output Layer}.
As illustrated in Figure~\ref{fig:chapter5_semantic_output_layer}, this variant changes the prediction head so that the hidden state is scored against semantic KC representations instead of being sent directly to a conventional classifier.
In the standard DKT setup, the hidden state is mapped directly to output logits over KCs by a learned linear layer.
In the semantic-output variant, the hidden state is first projected into the same space as the projected pretrained KC or question embeddings, and prediction is then based on a dot product between the projected hidden state and those stored semantic representations.
Writing \(h_t\) for the recurrent hidden state, the architecture is
\[
\hat{h}_t = W_h h_t + b_h, \qquad \bar{e}_k = W_{\mathrm{proj}} e_k, \qquad s_{t,k} = \hat{h}_t^\top \bar{e}_k,
\]
where \(e_k\) is the stored pretrained representation associated with KC \(k\), and \(W_{\mathrm{proj}}\) is the same projection family used on the input side.

Stored semantic targets are projected once into output space. One variant averages the two response-specific KC representations; another uses a single KC representation. This replaces the learned linear KC-output head with dot-product scores between the hidden state and stored semantic KC vectors. This reuse of input-embedding geometry at the output layer is analogous to \emph{weight tying} in NLP language models, but the analogy is limited to the shared geometry: this DKT head scores KCs rather than language-model tokens.

\FloatBarrier

\subsection{Full-Data Evaluation and Selection}

For RQ2.1, the \emph{full-data} setting uses all interactions in the trimmed text-linked dataset defined in Table~\ref{tab:chapter5_pb_data}. The pyKT preprocessing procedure assigns 20\% of the processed sequences to the test set and divides the remaining 80\% into five folds; fold 0 is used for validation and the other four folds for training. Each configuration combines a dataset, embedding model, representation pipeline, architecture variant, and any variant-specific settings. Across the full-data experiments, we evaluated approximately 1,065 distinct configurations; Supplementary Table~\ref{tab:app_ch5_full_data_configurations} summarizes the 21 experiment exports. This count refers to the RQ2.1 full-data modeling sweep, not to the RQ1 graph-alignment sweep. Within each experimental condition, the candidate with the highest validation AUC is selected; Table~\ref{tab:chapter5_results_combined} then reports its test AUC and test accuracy. Supplementary Tables~\ref{tab:app_ch5_pipeline_2009}--\ref{tab:app_ch5_pipeline_2017} report the corresponding per-dataset pipeline rows, with each pipeline represented by the configuration selected by validation AUC.

\subsection{Reduced-Data Protocol}

To test RQ2.2, we repeat the comparison with reduced datasets. This asks whether semantic representations provide a stronger inductive bias when behavioral data are limited.

The data-efficiency analysis is carried out on ASSISTments 2009, 2012, and 2017 by creating nested subsamples of each complete trimmed text-linked base dataset at \(10\%, 20\%, \dots, 90\%\) of that base dataset's row count. In this protocol, one row denotes one student interaction consisting of a KC and response. The sampling procedure is designed to preserve skill coverage: for each dataset, a mandatory set of rows is first selected so that every KC appearing in the complete trimmed base dataset appears at least once in the sampled dataset, and the remaining rows are then randomly permuted with a fixed seed. The resulting samples are nested, so the \(10\%\) sample is a subset of the \(20\%\) sample, the \(20\%\) sample is a subset of the \(30\%\) sample, and so on. Each sampled dataset is then processed with the same pyKT train/validation/test split procedure used above; the partitions are regenerated at each percentage rather than held fixed across percentages.

Two conditions are compared on these sampled datasets. The first is the baseline pyKT DKT model without pretrained semantic embeddings. The second initializes the model with a pretrained interaction matrix. The semantic representations are constructed from the complete trimmed text-linked question/KC resource. Across reduced-data percentages, the entire behavioral dataset is subsampled before splitting while the same external semantic matrices are reused. For ASSISTments 2009, all nine pipelines are evaluated; for ASSISTments 2012 and 2017, the reduced-data experiments use Pipeline~3.

We tune each sampled condition with a dataset-specific hyperparameter sweep managed in Weights \& Biases (W\&B). Final reporting follows the same validation-selection rule used in the full-data benchmark runs: for each condition, the run with the highest validation AUC is selected, and the selected run's test AUC and test accuracy are reported. For ASSISTments 2009, pipeline selection is nested inside that procedure, so the selected pretrained result at each sample size is the pipeline/run configuration with the highest validation AUC among the available semantic alternatives. Model selection is therefore based on validation performance rather than on the test set.

ASSISTments 2009 selects among nine pipelines, whereas ASSISTments 2012 and 2017 use fixed Pipeline~3. Because the search breadth differs by dataset, we treat the reduced-data patterns as descriptive rather than as controlled cross-dataset comparisons. The curves are based on one nested sampling path, so they do not show how results vary across alternative samples.

\subsection{KC-Supervised Encoder Analysis}

We trained dataset-specific BERT KC classifiers for all three datasets. The ASSISTments 2009 and 2012 encoders were used for downstream DKT initialization. We retained the ASSISTments 2017 classifier as a diagnostic because it had both a small filtered training set and weak held-out classification performance. These models differ from the general-purpose encoders in Table~\ref{tab:chapter5_embedding_models} because they are explicitly trained to predict KC labels from problem text rather than relying on general semantic similarity alone.

Fine-tuning is formulated as a multiclass text-classification problem in which the input is the raw question text and the target is the numeric KC index associated with that question. For each dataset, the dataset question--KC mapping (\texttt{questions\_with\_kc.csv}) supplies the question--KC pairs. Rows with missing text or skill identifiers are dropped, and KC classes with fewer than two examples are removed before the remaining examples are split independently for each dataset into stratified train, validation, and test partitions with a 70/15/15 ratio and random seed 42.

All fine-tuning experiments start from \texttt{bert-base-uncased}~\cite{devlin_bert_2019}. The model is instantiated as \texttt{BertForSequenceClassification} with the number of output labels set to the number of valid KC classes in the dataset and with hidden-state outputs enabled so that the encoder can later be reused for embedding extraction. Training uses AdamW with learning rate \(2 \times 10^{-5}\), weight decay \(0.01\), warmup over the first \(10\%\) of optimization steps, training batch size 16, evaluation batch size 32, maximum sequence length 128, and up to 10 epochs with early stopping based on validation loss using a patience of 3. Because the KC distributions are highly imbalanced, the loss is class-weighted cross-entropy with weights inversely proportional to class frequency in the training split. Gradient norms are clipped to 1.0, and the checkpoint with the lowest validation loss is kept.

After fine-tuning, the classification head is discarded and embeddings are extracted from the final-layer \texttt{[CLS]} representation, yielding a 768-dimensional vector for each question text. These per-question vectors are then passed through the same nine semantic-representation pipelines used for the general-purpose embedding models. For ASSISTments 2009 and 2012, the downstream tensors are constructed for the same trimmed text-linked KC sets used in the DKT comparisons and ordered by each dataset's KC index (\texttt{keyid2idx.\allowbreak json}).

The ASSISTments 2009 and 2012 tensors are loaded into DKT as pretrained interaction embeddings. Downstream BERT initialization is limited to ASSISTments 2009 and 2012, and no cross-dataset transfer is performed.

\section{Results}

\subsection{RQ1: Are the KC Relations Learned by DKT Semantic?}

\begin{figure}[!htbp]
    \centering
    \includegraphics[width=\linewidth,alt={Three heatmaps showing complete weighted DKT-derived KC relation matrices over the trimmed text-linked KC sets for ASSISTments 2009, 2012, and 2017.}]{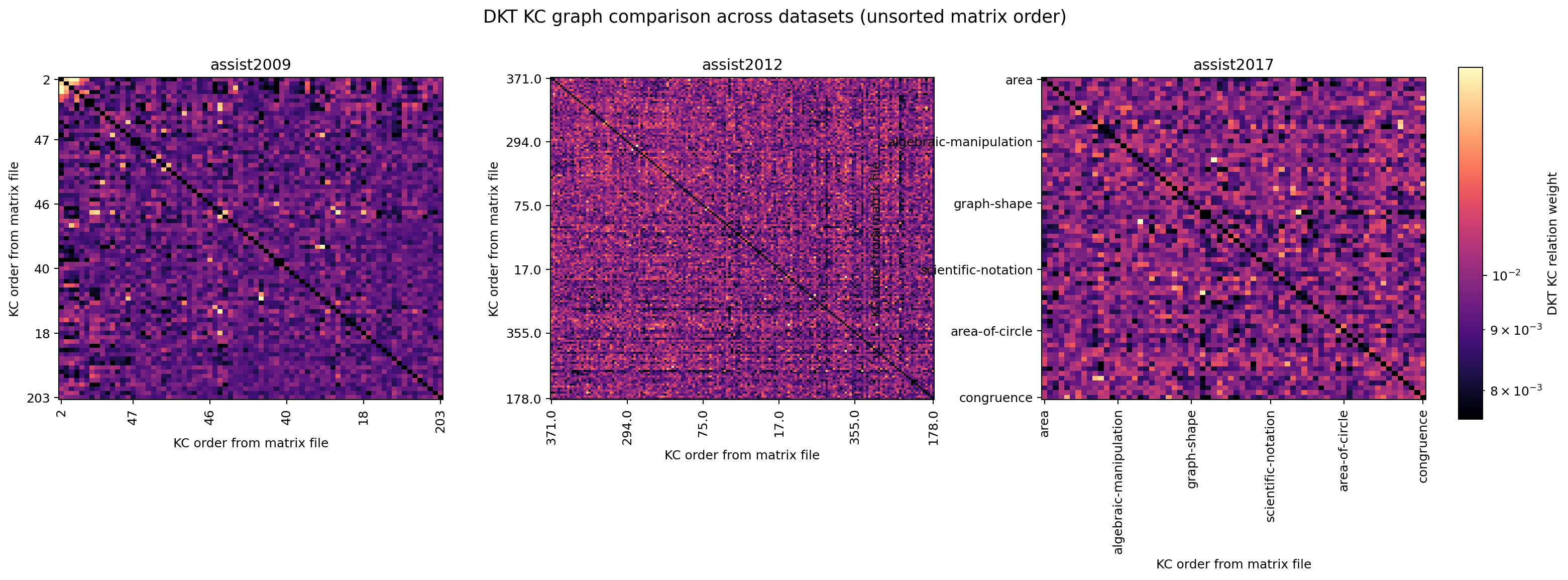}
    \caption{DKT-derived target matrices for RQ1. Each panel shows the symmetrized, zero-diagonal complete weighted relation matrix used for the RQ1 comparison over the corresponding trimmed text-linked KC set. KCs follow the numeric index order in that dataset's \texttt{keyid2idx.json}. Color normalization is panel-specific; the displayed colorbar corresponds to the ASSISTments 2017 panel.}
    \label{fig:chapter5_dkt_target_matrices}
    \Description{All three symmetric heatmaps have dark diagonals. The ASSISTments 2009 and 2017 panels contain scattered bright cells and pronounced horizontal or vertical bands, whereas the finer ASSISTments 2012 grid has a dense, mottled texture with less visible block structure.}
\end{figure}

The three DKT-derived target graphs serve as dataset-specific comparison targets (Figure~\ref{fig:chapter5_dkt_target_matrices}).

Their structures differ across datasets; the quantitative comparisons below test how well each dataset-specific semantic graph recovers them. ASSISTments 2012 is the dataset for which the embedding models achieve the highest, although still weak, agreement.

For RQ1, we compare DKT-derived and semantic graphs across the evaluated datasets, embedding models, pipelines, and architecture variants. Supplementary Table~\ref{tab:app_ch5_comparison_ledger} defines the comparison sets and aggregation units. Table~\ref{tab:chapter5_posthoc_best_by_dataset} reports the highest-Pearson configuration per dataset. ASSISTments 2012 is highest: \texttt{embed\_\allowbreak english\_\allowbreak v3.0} with Pipeline~3 reaches Pearson \(0.0894\) and top-10\% edge overlap \(0.1571\). ASSISTments 2009 is weaker, with \texttt{bert\_\allowbreak ft\_\allowbreak assist2009} and Pipeline~5 at Pearson \(0.0573\). ASSISTments 2017 is weakest, with \texttt{all\_\allowbreak MiniLM\_\allowbreak L6\_\allowbreak v2} and Pipeline~5 at Pearson \(0.0247\). All three semantic Pearson correlations in Table~\ref{tab:chapter5_posthoc_best_by_dataset} are positive but small, and none indicates strong recovery of the DKT-derived graph.

\begin{table}[!htbp]
    \centering
    \small
    \setlength{\tabcolsep}{5pt}
    \renewcommand{\arraystretch}{1.1}
    \caption{Highest-Pearson dataset-level alignment results between semantic KC representations and DKT-derived KC graphs. Each row is selected by Pearson correlation within that dataset's evaluated configurations. Spearman and top-10\% overlap are reported for the same Pearson-selected configuration. They are not selected separately by those metrics. Boldface marks the highest displayed value in each metric column.}
    \label{tab:chapter5_posthoc_best_by_dataset}
    \DataTableHeaderRows{1}
    \begin{tabular}{l l c c c c}
        \toprule
        Dataset & Embedding model & Pipeline & Pearson & Spearman & Top-10\% overlap \\
        \midrule
        ASSISTments 2009 & \texttt{bert\_ft\_assist2009} & 5 & 0.0573 & 0.0276 & 0.1151 \\
        ASSISTments 2012 & \texttt{embed\_english\_v3.0} & 3 & \textbf{0.0894} & \textbf{0.0782} & \textbf{0.1571} \\
        ASSISTments 2017 & \texttt{all\_MiniLM\_L6\_v2} & 5 & 0.0247 & 0.0114 & 0.1404 \\
        \bottomrule
    \end{tabular}
\end{table}

\FloatBarrier

As a non-text reference, training-sequence KC co-exposure aligns more strongly with the DKT-derived graph than the strongest semantic configuration on ASSISTments 2009 (Pearson \(0.1590\) versus \(0.0573\); Spearman \(0.0852\) versus \(0.0276\)) and ASSISTments 2012 (Pearson \(0.1457\) versus \(0.0894\); Spearman \(0.1433\) versus \(0.0782\)). The advantage is metric-dependent: top-10\% edge overlap favors the comparator on ASSISTments 2009 (\(0.2086\) versus \(0.1151\)) but is essentially tied on ASSISTments 2012 (\(0.1556\) versus \(0.1571\)). On ASSISTments 2017, the comparator is negative on both correlation metrics (Pearson \(-0.0322\), Spearman \(-0.0469\)). The comparator is a single unselected matrix per dataset, whereas the semantic values are maxima selected by Pearson across the alignment sweep.

Some embedding models align slightly better than others. Appendix Table~\ref{tab:app_ch5_posthoc_family_summary} reports embedding-\allowbreak model means for the five models shown, with each model aggregated over its evaluated dataset--pipeline configurations. Within these displayed rows, \texttt{bert\_\allowbreak ft\_\allowbreak assist2012} has the highest mean Pearson correlation (\(0.0321\)), followed by \texttt{text-\allowbreak embedding-\allowbreak 3-\allowbreak large}; even the highest mean remains small.

The highest-Pearson configurations use \texttt{bert\_\allowbreak ft\_\allowbreak assist2009} on ASSISTments 2009, \texttt{embed\_\allowbreak english\_\allowbreak v3.0} on ASSISTments 2012, and \texttt{all\_\allowbreak MiniLM\_\allowbreak L6\_\allowbreak v2} on ASSISTments 2017. These dataset-specific rankings do not identify one embedding model that consistently aligns with the relation structure extracted from DKT.

Appendix Table~\ref{tab:app_ch5_posthoc_pipeline_summary} reports pipeline means. Pipeline~5 has the highest mean Pearson correlation, followed by Pipelines~9 and~3, but none produces strong graph recovery.

Pipeline~5 has the highest alignment for ASSISTments 2009 and 2017 in the RQ1 alignment sweep, whereas Pipeline~3 is highest for ASSISTments 2012. Pipeline effects are therefore dataset-dependent and do not identify a universally successful semantic construction. Appendix Figure~\ref{fig:app_ch5_posthoc_pipeline_grid} shows that pipeline choice changes cosine-similarity structure without producing a common DKT-like organization.

ASSISTments 2012 contains the highest-Pearson configuration in the RQ1 alignment sweep. It also has the largest text-linked question pool in Table~\ref{tab:chapter5_pb_data}. A larger absolute pool of matched questions may help form more stable KC neighborhoods, but the observed alignment remains partial.

\begin{figure}[!htbp]
    \centering
    \begin{subfigure}[t]{0.48\linewidth}
        \centering
        \includegraphics[width=\linewidth]{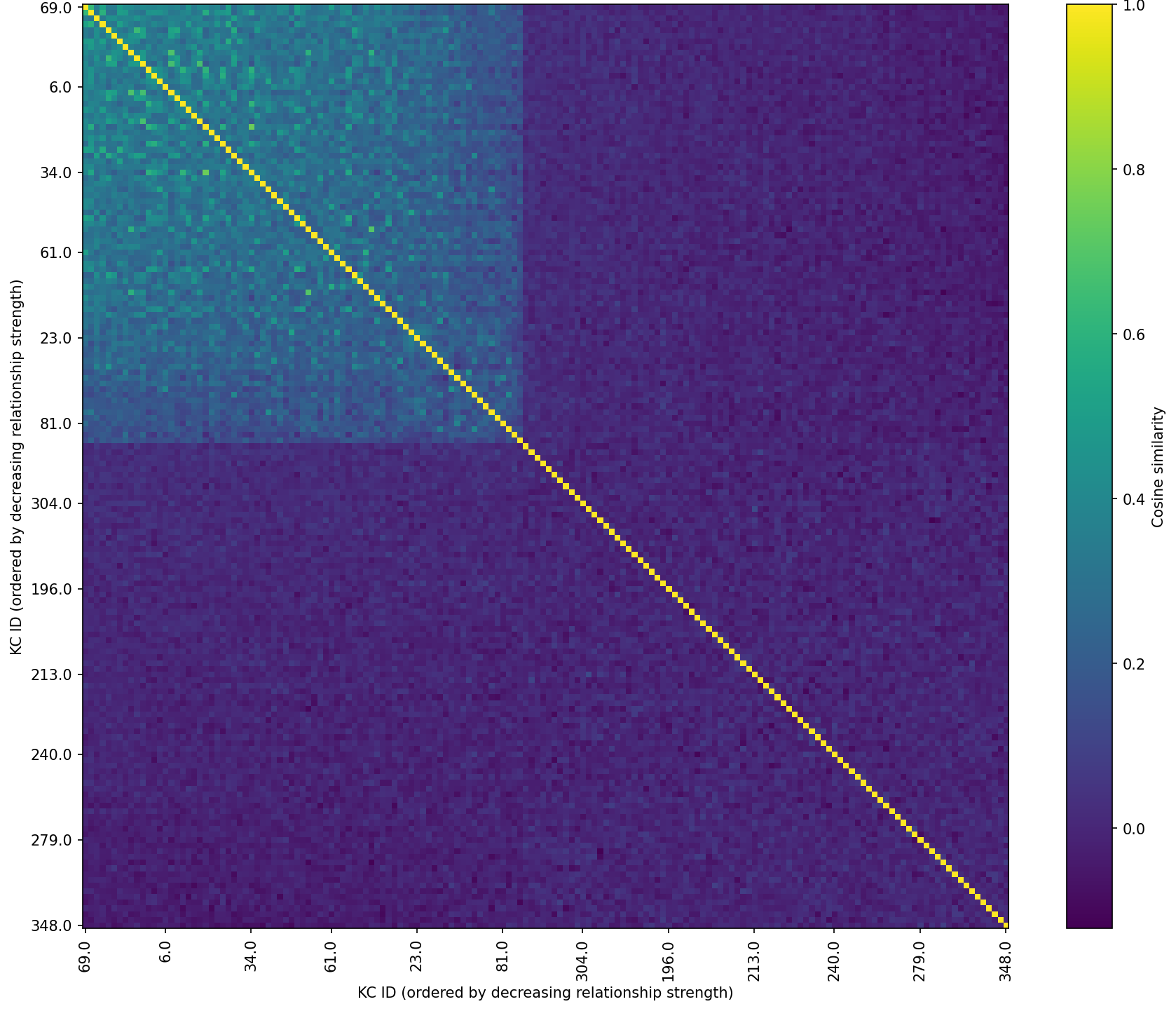}
        \caption{Embedding cosine-similarity matrix}
    \end{subfigure}\hfill
    \begin{subfigure}[t]{0.48\linewidth}
        \centering
        \includegraphics[width=\linewidth]{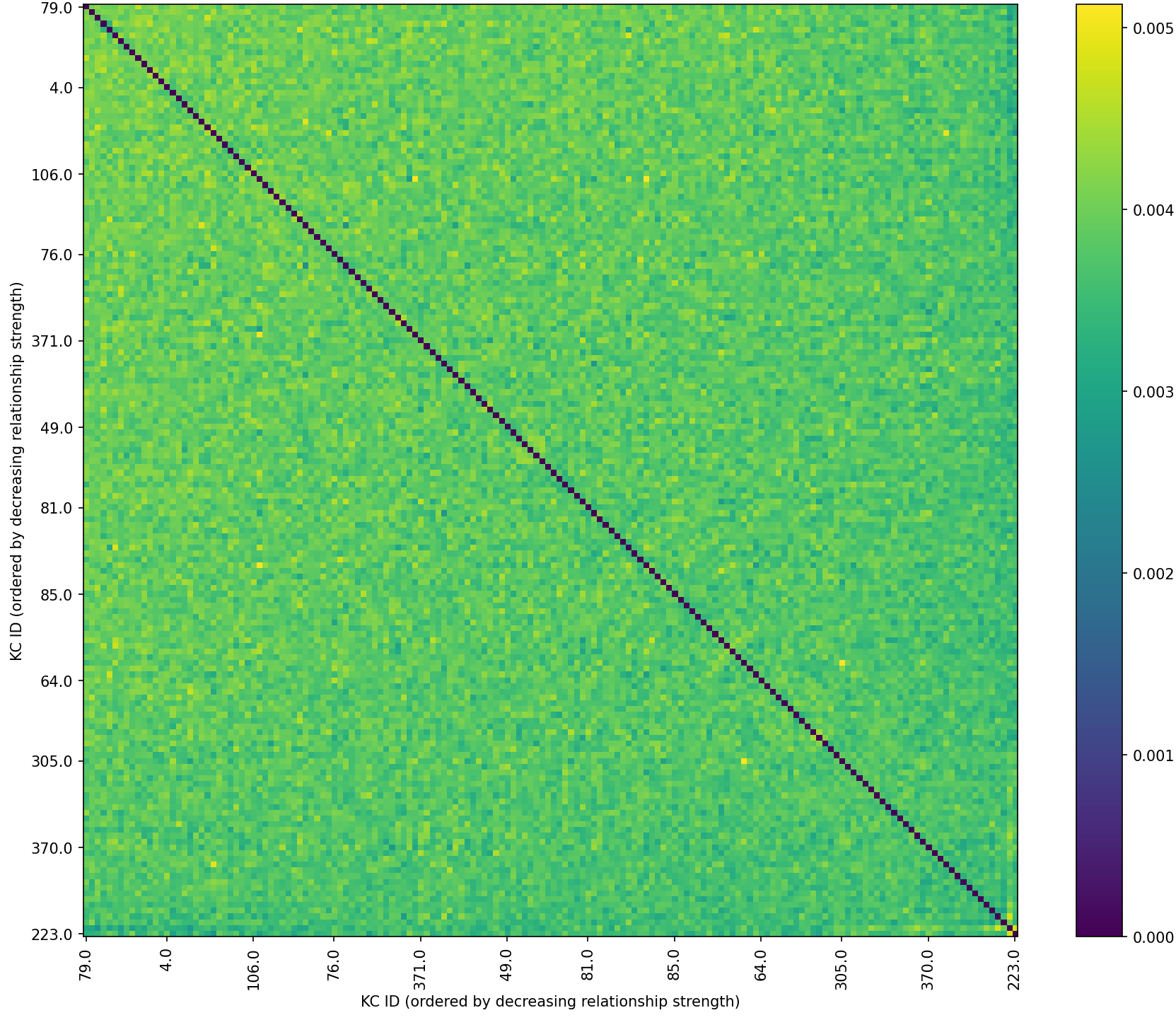}
        \caption{DKT-derived relation matrix (independently ordered)}
    \end{subfigure}
    \caption{Qualitative side-by-side view for the highest-Pearson ASSISTments 2012 configuration in the RQ1 alignment sweep (\texttt{embed\_english\_v3.0}, Pipeline~3). The panels use independent KC orders and color scales, so the figure compares overall structure rather than matched cells.}
    \label{fig:chapter5_posthoc_bestcase_heatmaps}
    \Description{The embedding matrix has a bright self-similarity diagonal and a large upper-left block of moderately high cosine similarity, with much lower similarities across the remaining rows and columns. The independently ordered DKT matrix has a dark zero diagonal and comparatively uniform off-diagonal weights, without a matching block pattern.}
\end{figure}

Figure~\ref{fig:chapter5_posthoc_bestcase_heatmaps} places the semantic and DKT-derived matrices for the highest-Pearson configuration in the RQ1 alignment sweep side by side: ASSISTments 2012 with \texttt{embed\_\allowbreak english\_\allowbreak v3.0} and Pipeline~3. Because the panels use independent KC orders, the figure is intended to show overall matrix structure rather than cellwise agreement.

\begin{figure}[!htbp]
    \centering
    \begin{subfigure}[t]{0.48\linewidth}
        \centering
        \begin{tikzpicture}
            \node[anchor=south west,inner sep=0] (agreement) at (0,0) {\includegraphics[height=0.16\textheight]{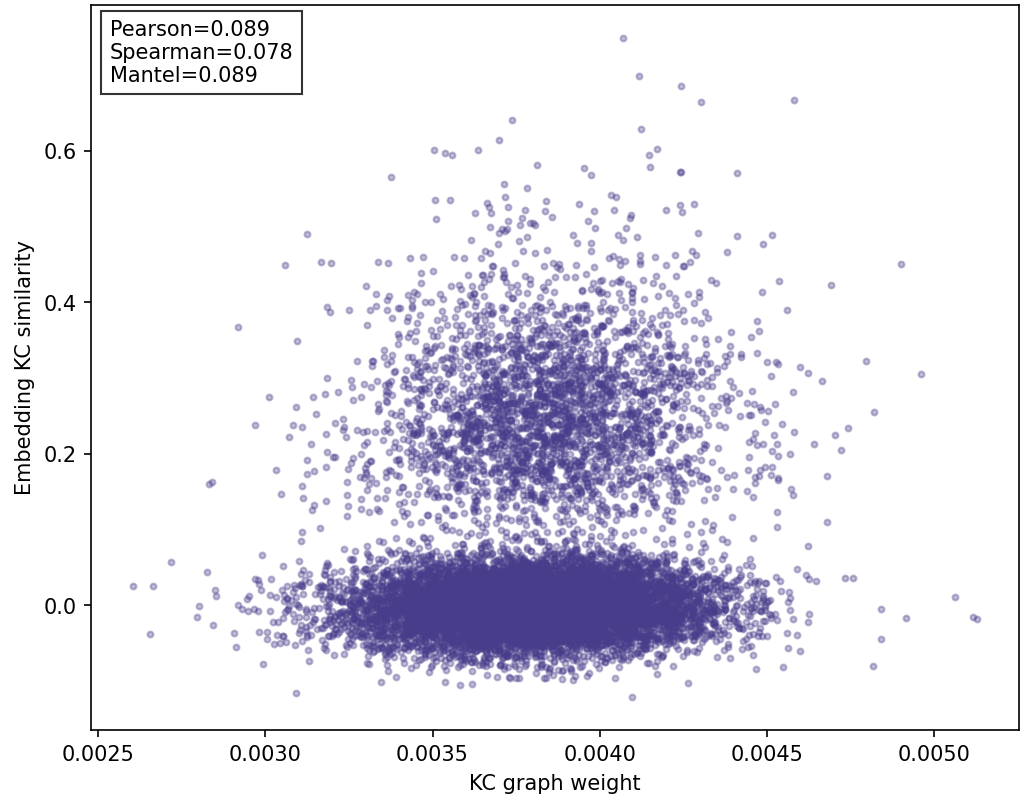}};
            \node[anchor=north west,fill=white,draw=black,inner sep=1.2pt,minimum width=15mm,minimum height=9mm,align=left,font=\fontsize{4}{4.6}\selectfont] at (agreement.north west) {Pearson=0.089\\Spearman=0.078};
        \end{tikzpicture}
        \caption{Pairwise relation agreement}
    \end{subfigure}\hfill
    \begin{subfigure}[t]{0.48\linewidth}
        \centering
        \includegraphics[height=0.16\textheight]{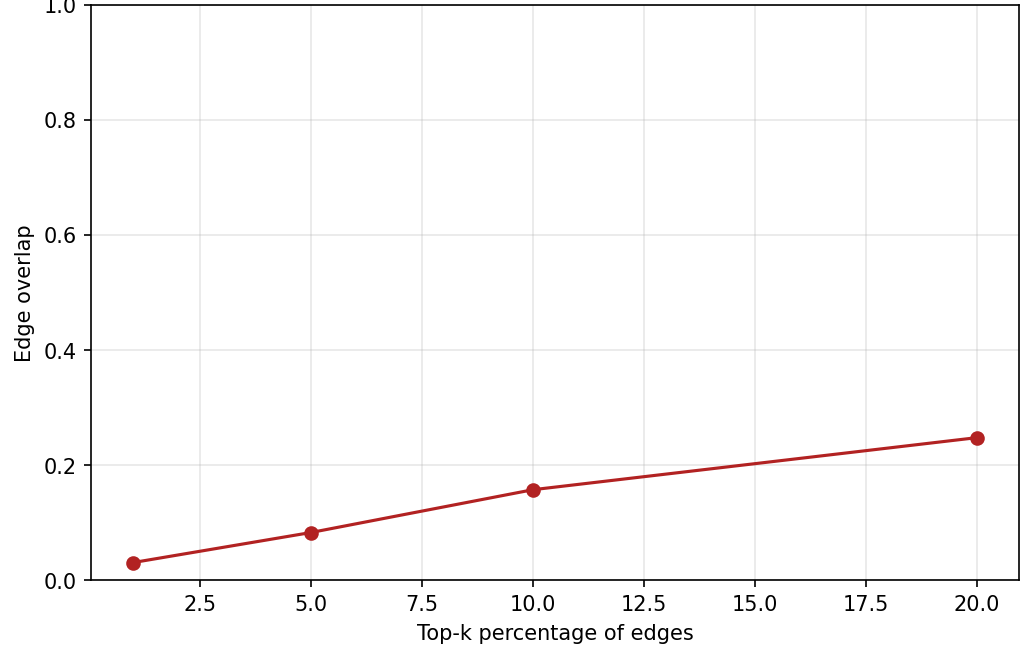}
        \caption{Top-\(k\) edge overlap}
    \end{subfigure}
\caption{Direct-agreement diagnostics for the Pearson-selected ASSISTments 2012 configuration in Table~\ref{tab:chapter5_posthoc_best_by_dataset}. The pairwise panel reports Pearson (0.089) and Spearman (0.078); the coefficients are descriptive.}
\label{fig:chapter5_posthoc_bestcase_alignment}
    \Description{The left scatterplot has a dense band of embedding similarities near zero and a broader cloud above it; the points remain diffuse, with only a weak positive trend as DKT graph weight increases. In the right panel, overlap between the highest-weight edge sets rises monotonically from the top 1 percent to the top 20 percent, reaching 0.1571 at 10 percent and about one-quarter at 20 percent.}
\end{figure}

Figure~\ref{fig:chapter5_posthoc_bestcase_alignment} shows the corresponding direct-agreement diagnostics. The displayed coefficients are Pearson \(0.089\) and Spearman \(0.078\). The trend is positive but diffuse, and the highest-weight DKT edges are only partially recovered by the semantic graph.

\FloatBarrier

These comparisons answer RQ1 directly: semantic KC representations align with DKT-derived KC graphs only weakly and inconsistently. Alignment is highest but partial in ASSISTments 2012, lowest in ASSISTments 2017, and intermediate in ASSISTments 2009. Embedding-model and pipeline choices shift the magnitude but not the conclusion: semantic similarity does not strongly recover the relation structure captured by the original DKT influence-score procedure.

\subsection{RQ2.1: Full-Data Modeling Performance}

The reported validation-selected semantic configurations do not exceed the baseline DKT test AUC on any dataset, although some produce localized gains in test accuracy; the AUC shortfalls range from 0.0011 to 0.0240, and these descriptive comparisons do not establish a consistent advantage for semantic initialization.

\noindent Table~\ref{tab:chapter5_results_combined} summarizes the validation-selected configuration for each full-data experimental condition in the study. Each row reports the associated test AUC and test accuracy, and the grouping labels follow the methodological variants introduced earlier. Supplementary Table~\ref{tab:app_ch5_full_data_configurations} summarizes the 21 experiment exports represented in the RQ2.1 full-data sweep.

\begin{table}[!htbp]
    \centering
    \small
    \setlength{\tabcolsep}{3pt}
    \renewcommand{\arraystretch}{0.98}
\caption{Combined full-data results across datasets. For each experimental condition, the reported row is the validation-selected configuration, together with its test AUC and test accuracy. Boldface indicates the highest value in each dataset block at the reported precision. Across the reported validation-selected semantic configurations, none exceeded the baseline test AUC on any dataset, although some configurations show localized accuracy gains. Pipe. denotes pipeline, SemOut the semantic-output condition, and Acc. accuracy.}
\label{tab:chapter5_results_combined}
    \DataTableHeaderRows{1}
    \resizebox{0.84\textwidth}{!}{%
    \begin{tabular}{lp{3.5cm}p{2.6cm}ccccc}
        \toprule
        Dataset & Variant & Embedding & Pipe. & SemOut & Test AUC & $\Delta$AUC & Test Acc. \\
        \midrule
        \multicolumn{8}{l}{\textbf{ASSISTments 2009}} \\
        2009 & Baseline DKT & -- & -- & No & \textbf{0.8582} & \textbf{+0.0000} & 0.7851 \\
        2009 & Pretrained Embedding Initialization & text-embedding-3-small & P1 & No & 0.8550 & -0.0031 & 0.7851 \\
        2009 & Pretrained Embedding Initialization + Regularization & text-embedding-3-small & P1 & No & 0.8567 & -0.0015 & 0.7865 \\
        2009 & Projection Bottleneck & text-embedding-3-large & P4 & No & 0.8552 & -0.0030 & \textbf{0.7870} \\
        2009 & Adaptive recurrent capacity & embed-v4.0 & P6 & No & 0.8342 & -0.0240 & 0.7708 \\
        2009 & Semantic Output Layer & text-embedding-3-small & P1 & Yes & 0.8522 & -0.0060 & 0.7811 \\
        2009 & Semantic Output Layer + Projection Bottleneck & bert\_\allowbreak ft\_\allowbreak assist2009 & P6 & Yes & 0.8349 & -0.0233 & 0.7727 \\
        \midrule
        \multicolumn{8}{l}{\textbf{ASSISTments 2012}} \\
        2012 & Baseline DKT & -- & -- & No & \textbf{0.7370} & \textbf{+0.0000} & \textbf{0.7437} \\
        2012 & Pretrained Embedding Initialization & text-embedding-3-large & P2 & No & 0.7265 & -0.0105 & 0.7344 \\
        2012 & Projection Bottleneck & text-embedding-3-large & P6 & No & 0.7351 & -0.0019 & 0.7373 \\
        2012 & Projection Bottleneck & embed-v4.0 & P6 & No & 0.7353 & -0.0016 & 0.7367 \\
        2012 & Adaptive recurrent capacity & text-embedding-3-large & P1 & No & 0.7317 & -0.0053 & 0.7365 \\
        2012 & Semantic Output Layer & text-embedding-3-large & P1 & Yes & 0.7356 & -0.0014 & 0.7376 \\
        2012 & Semantic Output Layer + Projection Bottleneck & text-embedding-3-large & P6 & Yes & 0.7358 & -0.0011 & 0.7371 \\
        \midrule
        \multicolumn{8}{l}{\textbf{ASSISTments 2017}} \\
        2017 & Baseline DKT & -- & -- & No & \textbf{0.7246} & \textbf{+0.0000} & \textbf{0.6889} \\
        2017 & Pretrained Embedding Initialization & text-embedding-3-small & P3 & No & 0.7207 & -0.0039 & 0.6811 \\
        2017 & Projection Bottleneck & text-embedding-3-small & P4 & No & 0.7196 & -0.0049 & 0.6808 \\
        2017 & Projection Bottleneck with Regularization & text-embedding-3-small & P6 & No & 0.7129 & -0.0116 & 0.6758 \\
        2017 & Adaptive recurrent capacity & text-embedding-3-small & P6 & No & 0.7197 & -0.0049 & 0.6803 \\
        2017 & Semantic Output Layer & text-embedding-3-large & P1 & Yes & 0.7195 & -0.0051 & 0.6785 \\
        2017 & Semantic Output Layer & text-embedding-3-large & P1 & Yes & 0.7193 & -0.0053 & 0.6798 \\
        \bottomrule
    \end{tabular}
    }
\end{table}

This result does not imply that language is irrelevant to KT. It shows that the semantic signal encoded by these pipelines, on its own, does not capture the dependencies that most strongly govern DKT predictions.

\subsubsection{Pipeline Comparison}

Appendix Tables~\ref{tab:app_ch5_pipeline_2009}--\ref{tab:app_ch5_pipeline_2017} compare the semantic variants, and Table~\ref{tab:chapter5_results_combined} provides the standard DKT baseline.

No pipeline is best across all three datasets. On ASSISTments 2009, Pipeline~1 achieved the highest semantic test AUC (\(0.8567\)) while Pipeline~4 achieved the highest semantic test accuracy (\(0.7870\)). On ASSISTments 2012, Pipelines~2, 3, and~6 tie for the highest reported semantic test AUC (\(0.7358\)), while Pipeline~7 achieved the highest semantic test accuracy (\(0.7382\)). On ASSISTments 2017, Pipeline~3 achieved the highest semantic test AUC (\(0.7207\)) and accuracy (\(0.6811\)). Pipeline results vary by dataset, but none of the reported semantic pipelines exceeds the baseline DKT test AUC on any dataset.

\FloatBarrier

\subsection{RQ2.2: Reduced-Data Modeling Performance}

For RQ2.2, the favorable ASSISTments 2017 trend suggests that semantic initialization may improve predictive performance in some lower-data settings, but the weak or unstable effects on ASSISTments 2009 and 2012 do not establish a general sample-efficiency advantage.

The reduced-data evaluation summarized in Table~\ref{tab:chapter5_data_efficiency_summary} tests whether semantic initialization improves DKT's predictive performance when less behavioral data are available. Model selection follows the protocol described earlier: for each sampled dataset, the representative run is selected by validation AUC, and the selected run's test AUC and test accuracy are reported. For ASSISTments 2009, this includes a pipeline-selection step across all nine pipelines at each sample size; for ASSISTments 2012 and 2017, the final sampled-data setup uses Pipeline 3 only.

\begin{table}[!htbp]
    \centering
    \small
    \setlength{\tabcolsep}{4pt}
    \renewcommand{\arraystretch}{1.1}
    \caption{Dataset-level summary of the data-efficiency experiments. Mean deltas are computed as pretrained minus baseline across the nine sampled percentages from \(10\%\) to \(90\%\) of each complete trimmed text-linked base dataset's interaction-row count; one row represents one student KC-response interaction. The Positive AUC splits column counts how often the selected pretrained run exceeds the selected baseline run on test AUC. The Largest \(\Delta\)AUC and Smallest \(\Delta\)AUC columns identify the sampled percentages with the largest and smallest pretrained-minus-baseline test-AUC differences, respectively. The Pipeline column identifies the pipeline used at the split with the largest difference.}
    \label{tab:chapter5_data_efficiency_summary}
    \DataTableHeaderRows{1}
    \begin{tabular}{lcccccc}
        \toprule
        Dataset & Mean $\Delta$AUC & Mean $\Delta$Acc. & \shortstack{Positive AUC\\splits} & \shortstack{Largest\\$\Delta$AUC} & Pipeline & \shortstack{Smallest\\$\Delta$AUC} \\
        \midrule
        ASSISTments 2009 & +0.0004 & +0.0000 & 5 / 9 & 40\% & P4 & 20\% \\
        ASSISTments 2012 & +0.0002 & -0.0028 & 6 / 9 & 80\% & P3 & 40\% \\
        ASSISTments 2017 & +0.0096 & +0.0020 & 8 / 9 & 30\% & P3 & 20\% \\
        \bottomrule
    \end{tabular}
\end{table}

\begin{figure}[!htbp]
    \centering
    \includegraphics[width=\linewidth]{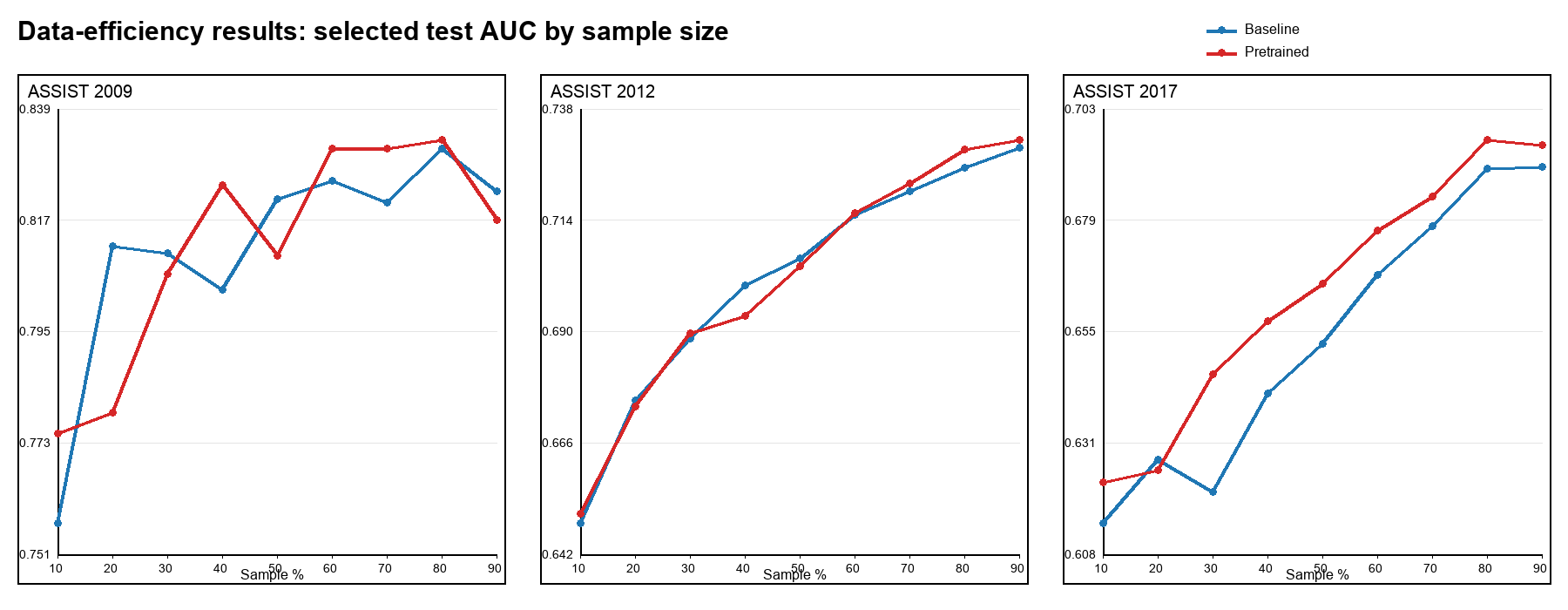}
\caption{Selected test AUC by sample size for the baseline and pretrained data-efficiency experiments. Each sample size is a percentage of the complete trimmed text-linked base dataset's interaction-row count, where one row represents one student KC-response interaction. Each curve uses the validation-selected run at each sampled percentage.}
\label{fig:chapter5_data_efficiency_auc}
\Description{Test AUC trends upward overall with sample size in all three panels. The ASSISTments 2009 baseline and pretrained curves cross repeatedly; the ASSISTments 2012 curves nearly overlap; and the ASSISTments 2017 pretrained curve lies above the baseline at eight of nine percentages, producing the clearest sustained advantage among the datasets.}
\end{figure}

Figure~\ref{fig:chapter5_data_efficiency_auc} reports the selected test AUC curves for the baseline and pretrained settings across sample sizes. The overall pattern is mixed. ASSISTments 2009 is unstable: the pretrained condition exceeds the baseline at five of the nine nested sample sizes, but the mean AUC gain is effectively zero and the selected pipeline changes with the sampling level. ASSISTments 2012 is similarly weak, with near-zero mean AUC change and a small decline in mean test accuracy. The most favorable dataset is ASSISTments 2017, for which the pretrained condition improves test AUC at eight of nine nested sample sizes and yields a mean gain of approximately 0.01 AUC.

\FloatBarrier

ASSISTments 2009 is the only reduced-data setting in which all nine pipelines were evaluated at each sample size. The detailed pipeline-by-pipeline view is provided in Appendix Figure~\ref{fig:app_ch5_data_efficiency_2009_pipelines}. No pipeline is best across all splits: the selected pipeline switches among P3, P4, P6, and P8, so the reduced-data ranking varies by split. By contrast, the reduced-data experiments for ASSISTments 2012 and 2017 use Pipeline~3 throughout, so those trends isolate the effect of semantic initialization more narrowly but do not test pipeline diversity in the same way.

\subsection{Mechanism Analysis: KC-Supervised Text Encoders}

The fine-tuned BERT experiments test a KC-label-supervised initialization strategy alongside the general-purpose embedding models used in the rest of the study. Instead of relying on general-purpose text similarity, these embeddings are derived from \texttt{bert-base-uncased} models fine-tuned to predict KC labels directly from question text. The representation should therefore be KC-discriminative for the dataset's labels. High classification performance shows that the encoder separates those labels, not that it captures semantic structure in general. The mechanism analysis asks whether that label alignment translates into better KT performance once the embeddings are introduced into DKT.

The fine-tuning results confirm that the encoder learns dataset-specific KC distinctions with highly variable success across datasets. Table~\ref{tab:chapter5_bert_classifier_results} reports test accuracies of \(0.9967\) and \(0.9795\) for ASSISTments 2009 and 2012, respectively, compared with \(0.1354\) for ASSISTments 2017. The filtered ASSISTments 2017 split contains 448 training examples and 96 test examples across 51 KC classes.

\begin{table}[!htbp]
    \centering
    \small
    \setlength{\tabcolsep}{5pt}
    \renewcommand{\arraystretch}{1.1}
\caption{KC-classification performance of the dataset-specific BERT classifiers after the stated filtering and split procedure. The \#KCs column reports the classifier's retained label count. The ASSISTments 2017 classifier is included as a diagnostic and was not used for downstream DKT initialization.}
\label{tab:chapter5_bert_classifier_results}
    \DataTableHeaderRows{1}
    \begin{tabular}{l r r r r r r}
        \toprule
        Dataset & \#KCs & Train & Test & Test Acc. & Macro F1 & Weighted F1 \\
        \midrule
        ASSISTments 2009 & 65 & 4297 & 921 & 0.9967 & 0.9665 & 0.9963 \\
        ASSISTments 2012 & 130 & 13894 & 2978 & 0.9795 & 0.8434 & 0.9772 \\
        ASSISTments 2017 & 51 & 448 & 96 & 0.1354 & 0.1095 & 0.0860 \\
        \bottomrule
    \end{tabular}
\end{table}

The ASSISTments 2017 classifier is included in Table~\ref{tab:chapter5_bert_classifier_results} as a diagnostic of this data constraint rather than as a downstream initialization condition. Its small filtered training set and weak held-out performance made it unsuitable for the fine-tuned BERT DKT comparison, so the downstream results in Table~\ref{tab:chapter5_bert_ft_results} are limited to ASSISTments 2009 and 2012.

\begin{table}[!htbp]
    \centering
    \small
    \setlength{\tabcolsep}{4pt}
    \renewcommand{\arraystretch}{1.1}
\caption{Matched downstream DKT comparison for the KC-supervised BERT encoders. The matched baseline is the no-embedding-initialization DKT baseline used for the same downstream comparison. For each dataset, the baseline and fine-tuned rows are selected by validation AUC within their respective comparison conditions. Boldface marks the highest displayed test value within each dataset and metric, including ties.}
\label{tab:chapter5_bert_ft_results}
    \DataTableHeaderRows{1}
    \begin{tabular}{l l l c c c c}
        \toprule
        Dataset & Condition & Selected pipeline & Test AUC & $\Delta$AUC & Test Acc. & $\Delta$Acc. \\
        \midrule
        ASSISTments 2009 & Matched baseline & -- & 0.8552 & -- & \textbf{0.7875} & -- \\
        ASSISTments 2009 & Fine-tuned BERT & P4 & \textbf{0.8581} & +0.0029 & \textbf{0.7875} & +0.0000 \\
        ASSISTments 2012 & Matched baseline & -- & 0.7363 & -- & \textbf{0.7384} & -- \\
        ASSISTments 2012 & Fine-tuned BERT & P4 & \textbf{0.7364} & +0.0001 & 0.7371 & -0.0013 \\
        \bottomrule
    \end{tabular}
\end{table}

\FloatBarrier

Table~\ref{tab:chapter5_bert_ft_results} shows that the KC-supervised BERT encoders produce small test-AUC gains over their matched baselines, but these gains do not alter the main conclusion of this study. On ASSISTments 2009, where Table~\ref{tab:chapter5_bert_classifier_results} shows near-perfect KC classification, the selected fine-tuned BERT configuration improves test AUC by 0.0029 over its matched baseline while leaving test accuracy unchanged. On ASSISTments 2012, the selected fine-tuned BERT configuration is essentially tied with its matched baseline on AUC and slightly lower on accuracy. In both datasets, the selected downstream configuration uses Pipeline~4. One possible mechanism is that constructing KC embeddings from a single representative question and then injecting response state through the stack-antonym scheme makes the KC-supervised representation more useful within these sweeps.

Both AUC differences are small descriptive values from validation-selected configurations, and the accuracy results do not show a corresponding improvement. Greater KC discriminability at the text-encoder level may therefore improve DKT's predictive performance in some configurations, but not consistently across the evaluated datasets and settings. The 2017 diagnostic shows what happens when the text-linked dataset is very small: the model struggles even to learn the KC classification problem itself, limiting the representations available for downstream KT.

\section{Discussion}

\subsection{Interpreting the Weak Graph Alignment}

The graph-alignment analysis directly answers RQ1, the paper's central question. If semantic KC representations encoded the same relational structure extracted from DKT's behavior by the influence-score procedure, then the semantic KC graphs would be expected to align strongly with the DKT-derived KC graphs. Instead, even the highest-Pearson configurations show only weak and inconsistent alignment, which is insufficient to support the claim that semantic KC geometry strongly recovers the relation structure captured by the original DKT influence-score procedure. Under this text-based test, DKT-derived relations are not primarily semantic.

Alignment is highest on ASSISTments 2012, lower on ASSISTments 2009, and lowest on ASSISTments 2017, but it remains weak in all three datasets. Embedding-model and pipeline choices improve alignment modestly, but no setting produces graph agreement strong enough to explain reliable downstream gains.

The co-exposure reference helps interpret the weak semantic alignment. On two of three datasets, a simple structure derived from the training sequences aligns more strongly with the DKT-derived matrix than the best text-semantic graph does. This suggests that the extracted graph contains a behavioral co-exposure component that text-derived geometry recovers less well. It does not establish the source of the DKT relation structure, and it is not a stability estimate.

This dataset ordering is also plausible given Table~\ref{tab:chapter5_pb_data}. ASSISTments 2012 has the largest absolute text-linked question pool, ASSISTments 2017 the smallest, and ASSISTments 2009 falls between them. That pattern does not prove that data volume alone determines graph alignment, but it does suggest that the absolute number of available question texts constrains how well semantic KC relations can be estimated.

ASSISTments 2017 has much less text per KC: 638 matched questions across 68 KCs, compared with thousands of matched questions across the modeled KC sets in ASSISTments 2009 and 2012. This makes the 2017 semantic KC representations higher-variance and plausibly contributes to both the weak alignment and the weak KC classifier. Text sparsity alone does not explain the full dataset pattern: the co-exposure reference uses no question text and is also weakest on ASSISTments 2017.

Dataset size alone is still not a sufficient explanation. Even the higher-alignment configurations on ASSISTments 2012 remain only partially aligned with the DKT graph. These results are consistent with a structural mismatch between text-derived semantic geometry and the relation structure captured by the DKT influence-score procedure, although the present experiments do not isolate that mismatch from all measurement and modeling choices.

The KC-supervised BERT configurations in the RQ1 alignment sweep likewise show only weak alignment. Different pipelines induce visibly different cosine-similarity patterns, so the construction choice is not trivial. However, the pipeline grid does not reveal any construction whose heatmaps consistently reorganize the KC space into something that resembles the DKT-derived graphs across datasets. These construction differences change the semantic geometry but do not produce strong recovery of the DKT-derived graphs.

The full-data experiments provide a complementary answer to RQ2.1. No reported semantic configuration selected by validation AUC exceeds the baseline DKT test AUC, although some configurations produce localized accuracy gains; pretrained semantic embeddings therefore do not reliably improve DKT's predictive performance in the full-data setting. This result is consistent with the weak graph alignment: if DKT primarily depended on relations captured by the evaluated semantic matrices, replacing random interaction embeddings with text-derived representations should have produced a more reliable benefit. However, predictive performance also depends on representation construction, optimization, and model architecture, so the initialization experiments do not independently establish that DKT relations are not primarily semantic.

\subsection{What the Limited Favorable Signals Mean}

The reduced-data experiments provide a mixed answer to RQ2.2. If pretrained semantic embeddings provided a strong inductive bias for KT, their advantage would be expected to emerge more clearly when less behavioral data are available. The favorable ASSISTments 2017 trend includes positive AUC changes at eight of nine nested sample sizes and an approximately 0.01 mean AUC improvement, whereas the effects on ASSISTments 2009 and 2012 are weak or unstable. This pattern does not establish a general sample-efficiency advantage. The study does not isolate the cause of this dataset contrast; the smaller absolute text-linked question pool on ASSISTments 2017 and the larger but potentially more heterogeneous KC space on ASSISTments 2012 may contribute, but neither explanation is established by these experiments.

The KC-supervised BERT analysis tests whether dataset-specific supervision changes the full-data result. Dataset-specific supervision makes semantic initialization more competitive, especially on ASSISTments 2009, yet the gains remain small and inconsistent and fail to produce a decisive full-data advantage. The mechanism analysis therefore qualifies but does not overturn the answer to RQ2.1.

These results limit direct transfer from text-only NLP representations to educational modeling. In KT, the relevant structure may depend on response behavior, temporal order, latent prerequisite dependencies, psychometric difficulty, and dataset-specific instructional organization. These factors may correlate with semantics, but they need not be reducible to semantics. The present results therefore suggest that KC relations in KT may occupy a latent learning space: a representation of how different KCs are organized in the learner's knowledge state, only weakly aligned with the semantic geometry induced by text embeddings. A prerequisite relation can be directional, whereas cosine similarity is symmetric; semantic proximity therefore cannot represent every property of a directed learning relation.

\subsection{Broader Implications for Knowledge Tracing}

Under the evaluated configurations, simply substituting semantic embeddings into the KT input representation does not reliably improve DKT's predictive performance. Future approaches may need to combine text with behavioral, temporal, curricular, structural, or psychometric signals rather than only modify the input representation.

\subsection{Limitations}

This study uses three ASSISTments datasets, and question-text coverage is incomplete and uneven, as shown in Table~\ref{tab:chapter5_pb_data}. The study is restricted to the text-linked subset of each dataset, so the conclusions apply to questions and KCs with matched Problem Bodies coverage rather than to every interaction or KC in the original full logs. This scope is especially restrictive for ASSISTments 2017.

The reported comparisons are also conditional on the evaluated embedding models and nine representation pipelines. These choices include KC-supervised BERT encoders and several recurrent-model variants, but they do not exhaust all possible ways of representing semantic structure or integrating language into KT. RQ2 therefore tests these particular ways of introducing semantic representations into DKT. Those conditions also change input dimensionality and recurrent capacity, so they are not a pure test of semantic content alone. A dimension-matched random control would help separate these factors in future work.

Several reported AUC and accuracy differences are small descriptive values from validation-selected configurations. The study does not quantify run-to-run variation for those selected full-data runs or variability across alternative reduced-data samples, so the favorable values should not be read as statistically established effects.

The reported analyses follow the processed single-KC interaction representation used in the DKT comparisons, although ASSISTments interactions can involve one or more KCs. Richer treatments of multi-KC structure may change how semantic and behavioral relations are represented and remain outside the scope of the present experiments.

The reduced-data test-AUC pattern is favorable on ASSISTments 2017 but weak or unstable on ASSISTments 2009 and 2012. This dataset dependence does not support a general sample-efficiency claim.

Finally, the graph-alignment analysis depends on the complete weighted DKT relation matrices over the trimmed text-linked KC sets produced by the original DKT influence-score extraction procedure and on the evaluated model and training setup. The comparison symmetrizes the directional DKT relations before comparing them with cosine similarity. The alignment findings therefore apply to this graph-extraction and comparison procedure, not to every possible notion of KT structure or conceptual relatedness.

We do not estimate how stable the extracted DKT relation matrices are across retraining runs, so the observed alignment magnitudes are not calibrated against the reproducibility of the DKT-derived graph itself. The co-exposure reference depends on the composition and length of the processed training sequences and collapses response order and correctness. It is a descriptive behavioral reference, not a stability ceiling for the DKT-derived graph.

\section{Conclusion}

We asked: Are the KC relations learned by DKT semantic? DKT models relations among KCs because an interaction with one KC can affect predictions for other KCs. In the operational sense tested here, the relations are not semantic: the tested text-derived semantic structures do not strongly recover the DKT-derived relation structure.

DKT configurations initialized with text-derived semantic KC representations do not reliably outperform the standard DKT baseline. The reduced-data test-AUC pattern is favorable on ASSISTments 2017, and KC-supervised BERT produces small AUC gains on ASSISTments 2009 and 2012, but these findings do not establish a consistent predictive advantage across the evaluated datasets and settings. We interpret these results as consistent with a learning-relevant latent space that is only weakly aligned with semantic proximity among questions and KCs.

\FloatBarrier
\section*{Declaration of Generative AI Software Tools in the Writing Process}

AI-based writing tools were used to assist with copyediting and language refinement; all research, analysis, interpretation, and conclusions are the author's own.

\bibliographystyle{acmtrans}
\bibliography{references}

\clearpage
\appendix
\section{Supplementary Material}

\subsection{Comparison Sets and Aggregation}
\label{sec:app_ch5_comparison_ledger}

\begin{table}[!htbp]
    \centering
    \scriptsize
    \setlength{\tabcolsep}{4pt}
    \renewcommand{\arraystretch}{1.15}
    \caption{Comparison sets and aggregation units for the graph-alignment results. Because coverage varies across the evaluated configurations, the aggregate means are descriptive summaries rather than controlled effects of changing only one configuration dimension.}
    \label{tab:app_ch5_comparison_ledger}
    \DataTableHeaderRows{1}
    \begin{tabularx}{\textwidth}{>{\raggedright\arraybackslash}p{0.22\textwidth}>{\raggedright\arraybackslash}X}
        \toprule
        Term or display & Scope and aggregation unit \\
        \midrule
        Configuration & One dataset-specific combination of embedding model, representation pipeline, and architecture variant. \\
        RQ1 alignment sweep & The full set of evaluated dataset--embedding-model--representation-pipeline--architecture-variant combinations; coverage varies because not every model--pipeline--architecture combination is present in every dataset. \\
        RQ2.1 full-data sweep & Approximately 1,065 distinct evaluated full-data configurations represented across the 21 experiment exports summarized in Table~\ref{tab:app_ch5_full_data_configurations}. \\
        Displayed embedding-model summary & The five model means reported in Table~\ref{tab:app_ch5_posthoc_family_summary}; this display does not define which configurations contribute to the pipeline aggregate. \\
        Embedding-model mean & Reported mean across each displayed model's evaluated dataset--pipeline configurations. \\
        Pipeline mean & Reported across-dataset aggregate for that pipeline. \\
        Table~\ref{tab:chapter5_posthoc_best_by_dataset} & One configuration per dataset, selected by Pearson correlation from the RQ1 alignment sweep. Spearman and top-10\% overlap are conditional diagnostics for those rows. \\
        Tables~\ref{tab:app_ch5_posthoc_family_summary}--\ref{tab:app_ch5_posthoc_pipeline_summary} & Reported aggregate means, not configuration-level maxima or a complete RQ1-sweep distribution. \\
        \bottomrule
    \end{tabularx}
\end{table}

\subsection{Detailed Pipeline Comparisons}
\label{sec:appendix_ch5_pipelines}

Tables~\ref{tab:app_ch5_pipeline_2009}--\ref{tab:app_ch5_pipeline_2017} report the per-dataset pipeline results. Across the three datasets, the highest-AUC semantic pipeline varies, and no semantic pipeline consistently outperforms the standard DKT baseline in test AUC.

\begin{table}[!htbp]
    \centering
    \small
    \setlength{\tabcolsep}{6pt}
    \renewcommand{\arraystretch}{1.1}
    \caption{Detailed pipeline comparison for ASSISTments 2009, with each pipeline represented by its validation-selected configuration. Boldface marks the highest displayed test value in each metric column. Pipeline~1 achieved the highest semantic test AUC, while Pipeline~4 achieved the highest semantic test accuracy.}
    \label{tab:app_ch5_pipeline_2009}
    \DataTableHeaderRows{1}
    \begin{tabular}{l l c c}
        \toprule
        Pipeline & Selected architecture & Test AUC & Test Acc. \\
        \midrule
        P1 & Pretrained Embedding Initialization + Regularization & \textbf{0.8567} & 0.7865 \\
        P2 & Semantic Output Layer + Projection Bottleneck & 0.8321 & 0.7702 \\
        P3 & Projection Bottleneck & 0.8542 & 0.7845 \\
        P4 & Projection Bottleneck & 0.8552 & \textbf{0.7870} \\
        P5 & Projection Bottleneck & 0.8535 & 0.7847 \\
        P6 & Semantic Output Layer + Projection Bottleneck & 0.8349 & 0.7727 \\
        P7 & Projection Bottleneck & 0.8536 & 0.7843 \\
        P8 & Projection Bottleneck & 0.8535 & 0.7852 \\
        P9 & Projection Bottleneck & 0.8542 & 0.7869 \\
        \bottomrule
    \end{tabular}
\end{table}

\begin{table}[!htbp]
    \centering
    \small
    \setlength{\tabcolsep}{6pt}
    \renewcommand{\arraystretch}{1.1}
    \caption{Detailed pipeline comparison for ASSISTments 2012, with each pipeline represented by its validation-selected configuration. Boldface marks the highest displayed test value in each metric column, including ties. Pipelines~2, 3, and~6 tie for the highest semantic test AUC, while Pipeline~7 achieved the highest semantic test accuracy.}
    \label{tab:app_ch5_pipeline_2012}
    \DataTableHeaderRows{1}
    \begin{tabular}{l l c c}
        \toprule
        Pipeline & Selected architecture & Test AUC & Test Acc. \\
        \midrule
        P1 & Semantic Output Layer & 0.7356 & 0.7376 \\
        P2 & Semantic Output Layer + Projection Bottleneck & \textbf{0.7358} & 0.7377 \\
        P3 & Semantic Output Layer + Projection Bottleneck & \textbf{0.7358} & 0.7374 \\
        P4 & Semantic Output Layer + Projection Bottleneck & 0.7354 & 0.7376 \\
        P5 & Semantic Output Layer + Projection Bottleneck & 0.7353 & 0.7374 \\
        P6 & Semantic Output Layer + Projection Bottleneck & \textbf{0.7358} & 0.7371 \\
        P7 & Semantic Output Layer + Projection Bottleneck & 0.7355 & \textbf{0.7382} \\
        P8 & Semantic Output Layer + Projection Bottleneck & 0.7356 & 0.7372 \\
        P9 & Semantic Output Layer + Projection Bottleneck & 0.7350 & 0.7367 \\
        \bottomrule
    \end{tabular}
\end{table}

\begin{table}[!htbp]
    \centering
    \small
    \setlength{\tabcolsep}{6pt}
    \renewcommand{\arraystretch}{1.1}
    \caption{Detailed pipeline comparison for ASSISTments 2017, with each pipeline represented by its validation-selected configuration. Boldface marks the highest displayed test value in each metric column. Pipeline~3 achieved the highest semantic test AUC and test accuracy.}
    \label{tab:app_ch5_pipeline_2017}
    \DataTableHeaderRows{1}
    \begin{tabular}{l l c c}
        \toprule
        Pipeline & Selected architecture & Test AUC & Test Acc. \\
        \midrule
        P1 & Pretrained Embedding Initialization & 0.7205 & 0.6806 \\
        P2 & Semantic Output Layer & 0.7171 & 0.6757 \\
        P3 & Pretrained Embedding Initialization & \textbf{0.7207} & \textbf{0.6811} \\
        P4 & Projection Bottleneck & 0.7196 & 0.6808 \\
        P5 & Projection Bottleneck & 0.7137 & 0.6774 \\
        P6 & Adaptive recurrent capacity & 0.7197 & 0.6803 \\
        P7 & Adaptive recurrent capacity & 0.7141 & 0.6788 \\
        P8 & Projection Bottleneck with Regularization & 0.7009 & 0.6719 \\
        P9 & Projection Bottleneck with Regularization & 0.6997 & 0.6711 \\
        \bottomrule
    \end{tabular}
\end{table}

\FloatBarrier

\subsection{Full-Data Sweep Coverage}
\label{sec:appendix_ch5_full_data_configurations}

Table~\ref{tab:app_ch5_full_data_configurations} summarizes the 21 experiment exports that represent approximately 1,065 distinct evaluated configurations in the RQ2.1 full-data sweep. Together, the exports contain 2,324 records, including 1,605 finished runs with usable validation and test AUC values. Repeated logged configurations may appear, so these record totals are not counts of distinct configurations.

\begin{center}
    \footnotesize
    \setlength{\tabcolsep}{2.3pt}
    \renewcommand{\arraystretch}{0.90}
    \captionof{table}{Experiment-export coverage for the RQ2.1 full-data sweep. The Exported column gives the number of records in each export; the Usable column gives the number of finished runs with usable validation and test AUC values. Selection is by validation AUC. Model labels: 3-small = \texttt{text-embedding-3-small}; 3-large = \texttt{text-embedding-3-large}; bert-ft-2009 = \texttt{bert\_ft\_assist2009}. Repeated logged configurations may appear in an export.}
    \label{tab:app_ch5_full_data_configurations}
    \DataTableHeaderRows{1}
    \begin{tabularx}{\textwidth}{@{}l c >{\raggedright\arraybackslash}X >{\raggedright\arraybackslash}p{0.19\textwidth} r r c@{}}
        \toprule
        ID & Dataset & Condition / variant & Embedding model & Exported & Usable & Selection \\
        \midrule
        S01 & 2009 & Baseline DKT & -- & 159 & 139 & Val. AUC \\
        S02 & 2009 & Pretrained initialization & 3-small & 40 & 40 & Val. AUC \\
        S03 & 2009 & Initialization + regularization & 3-small & 10 & 10 & Val. AUC \\
        S04 & 2009 & Projection bottleneck & 3-large & 27 & 27 & Val. AUC \\
        S05 & 2009 & Adaptive recurrent capacity & \path{embed-v4.0} & 27 & 27 & Val. AUC \\
        S06 & 2009 & Semantic output & 3-small & 45 & 44 & Val. AUC \\
        S07 & 2009 & Semantic output + projection & bert-ft-2009 & 27 & 27 & Val. AUC \\
        \addlinespace[1pt]
        S08 & 2012 & Baseline DKT & -- & 20 & 20 & Val. AUC \\
        S09 & 2012 & Pretrained initialization & 3-large & 25 & 24 & Val. AUC \\
        S10 & 2012 & Projection bottleneck & 3-large & 36 & 36 & Val. AUC \\
        S11 & 2012 & Projection bottleneck & \path{embed-v4.0} & 36 & 36 & Val. AUC \\
        S12 & 2012 & Adaptive recurrent capacity & 3-large & 80 & 79 & Val. AUC \\
        S13 & 2012 & Semantic output & 3-large & 34 & 33 & Val. AUC \\
        S14 & 2012 & Semantic output + projection & 3-large & 124 & 121 & Val. AUC \\
        \addlinespace[1pt]
        S15 & 2017 & Baseline DKT & -- & 100 & 100 & Val. AUC \\
        S16 & 2017 & Pretrained initialization & 3-small & 796 & 113 & Val. AUC \\
        S17 & 2017 & Projection bottleneck & 3-small & 166 & 166 & Val. AUC \\
        S18 & 2017 & Projection + regularization & 3-small & 60 & 60 & Val. AUC \\
        S19 & 2017 & Adaptive recurrent capacity & 3-small & 143 & 137 & Val. AUC \\
        S20 & 2017 & Semantic output, export 1 & 3-large & 173 & 171 & Val. AUC \\
        S21 & 2017 & Semantic output, export 2 & 3-large & 196 & 195 & Val. AUC \\
        \bottomrule
    \end{tabularx}
\end{center}

\FloatBarrier

\subsection{Additional Data-Efficiency Diagnostic}
\label{sec:appendix_ch5_data_efficiency}

\begin{figure}[!htbp]
    \centering
    \includegraphics[width=0.92\linewidth]{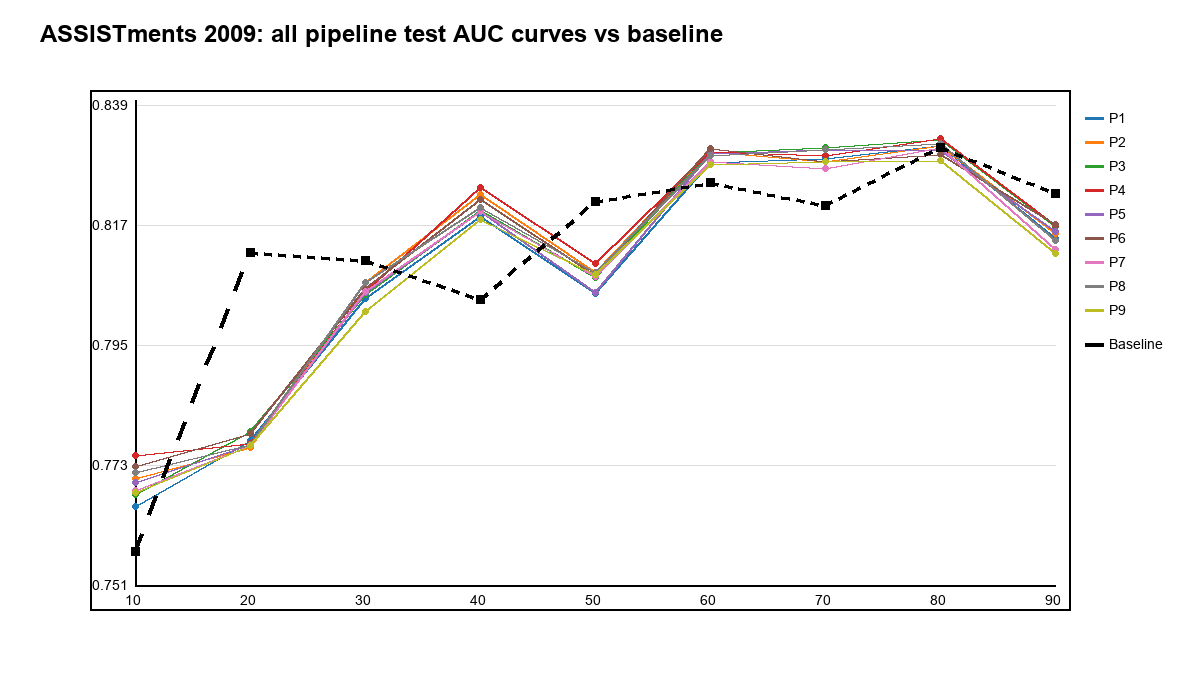}
    \caption{ASSISTments 2009 data-efficiency comparison across all nine pipelines and the baseline. Each sample size is a percentage of the complete trimmed text-linked base dataset's interaction-row count, where one row represents one student KC-response interaction. Each pipeline curve shows the validation-selected configuration for that pipeline at each sampled percentage.}
    \Description{Nine colored pipeline curves remain tightly clustered and cross one another as sample size increases, so no pipeline stays highest. The dashed baseline alternates relative to this cluster: it is lower at 10 and 40 percent, higher at 20, 30, and 50 percent, lower again at 60 and 70 percent, nearly tied with the cluster at 80 percent, and higher at 90 percent.}
    \label{fig:app_ch5_data_efficiency_2009_pipelines}
\end{figure}

Figure~\ref{fig:app_ch5_data_efficiency_2009_pipelines} gives the full ASSISTments 2009 pipeline-by-pipeline reduced-data comparison. ASSISTments 2009 is the only dataset in which all nine pipelines were compared at every sample size. The selected pipeline changes across splits, so the result does not favor one semantic construction.

\FloatBarrier

\subsection{Additional KC-Graph Alignment Diagnostics}
\label{sec:appendix_ch5_posthoc}

Tables~\ref{tab:app_ch5_posthoc_family_summary}--\ref{tab:app_ch5_posthoc_pipeline_summary} report embedding-model and pipeline means for the KC-graph alignment analysis. The rankings vary, but no displayed aggregate strongly recovers the DKT-derived graph. Table~\ref{tab:app_ch5_comparison_ledger} defines the comparison sets and aggregation units.

\begin{table}[!htbp]
    \centering
    \small
    \setlength{\tabcolsep}{5pt}
    \renewcommand{\arraystretch}{1.1}
    \caption{Displayed embedding-model means. Models are ordered by mean Pearson correlation across each model's evaluated dataset--pipeline configurations. Boldface marks the highest displayed value in each metric column, including ties.}
    \label{tab:app_ch5_posthoc_family_summary}
    \DataTableHeaderRows{1}
    \begin{tabular}{l c c c}
        \toprule
        Embedding model & Pearson & Spearman & Top-10\% overlap \\
        \midrule
        \texttt{bert\_ft\_assist2012} & \textbf{0.0321} & \textbf{0.0260} & \textbf{0.1304} \\
        \texttt{text-embedding-3-large} & 0.0297 & \textbf{0.0260} & 0.1268 \\
        \texttt{bert\_ft\_assist2009} & 0.0217 & 0.0118 & 0.0999 \\
        \texttt{all-MiniLM-L12-v2} & 0.0205 & 0.0178 & 0.1260 \\
        \texttt{embed-v4.0} & 0.0187 & 0.0185 & 0.1229 \\
        \bottomrule
    \end{tabular}
\end{table}

\begin{table}[!htbp]
    \centering
    \small
    \setlength{\tabcolsep}{5pt}
    \renewcommand{\arraystretch}{1.1}
    \caption{Reported pipeline summary across datasets. Pipelines are ordered by mean Pearson correlation. Boldface marks the highest displayed value in each metric column.}
    \label{tab:app_ch5_posthoc_pipeline_summary}
    \DataTableHeaderRows{1}
    \begin{tabular}{c c c c}
        \toprule
        Pipeline & Pearson & Spearman & Top-10\% overlap \\
        \midrule
        5 & \textbf{0.0243} & 0.0233 & 0.1229 \\
        9 & 0.0238 & 0.0223 & 0.1214 \\
        3 & 0.0227 & 0.0191 & 0.1228 \\
        7 & 0.0189 & 0.0165 & 0.1208 \\
        1 & 0.0175 & 0.0142 & 0.1216 \\
        6 & 0.0167 & 0.0237 & 0.1231 \\
        4 & 0.0159 & \textbf{0.0243} & \textbf{0.1264} \\
        8 & 0.0138 & 0.0163 & 0.1144 \\
        2 & 0.0114 & 0.0104 & 0.1229 \\
        \bottomrule
    \end{tabular}
\end{table}

\FloatBarrier

\begin{figure}[!htbp]
    \centering
    \includegraphics[width=\linewidth]{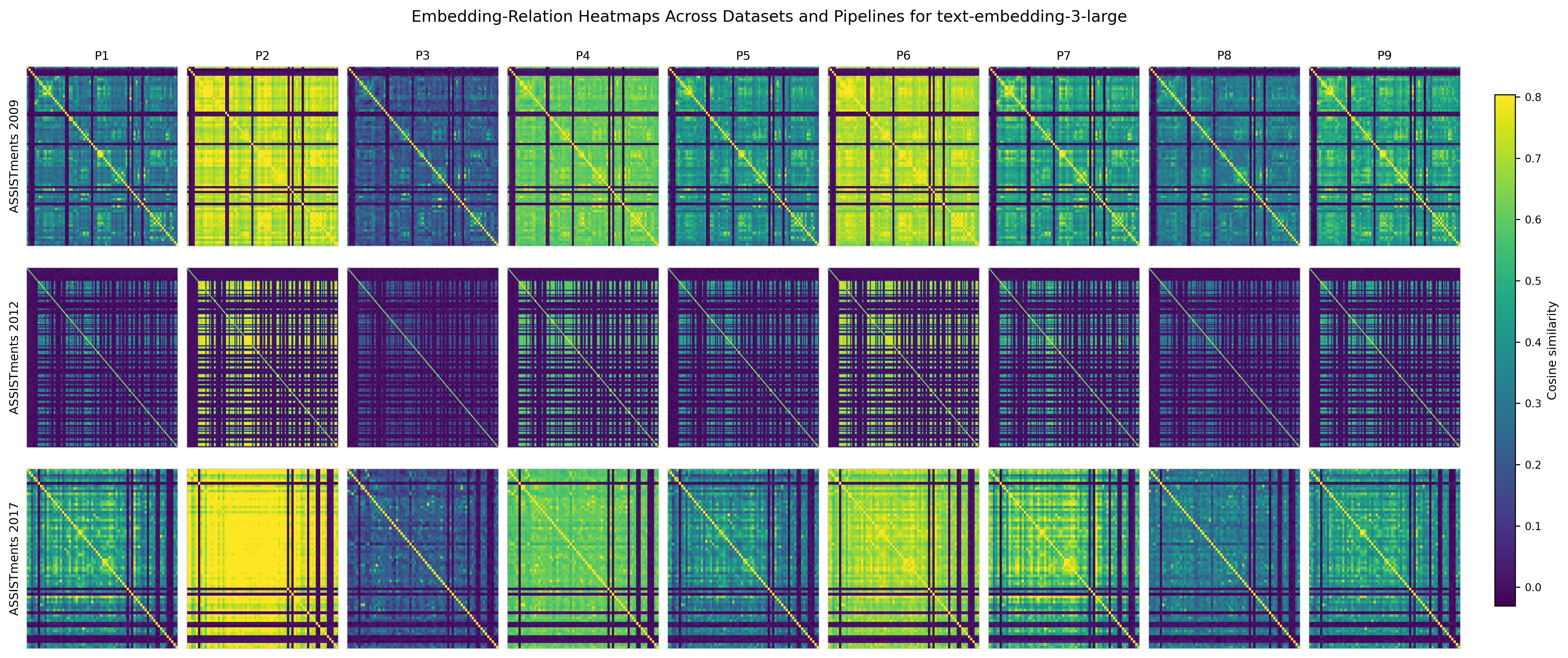}
    \caption{Pipeline-induced embedding-relation heatmaps for \texttt{text-embedding-3-large}.}
    \Description{A three-by-nine grid has one row for each ASSISTments dataset and one column for Pipelines 1--9. Each symmetric cosine-similarity heatmap has a bright diagonal and horizontal or vertical bands of high and low similarity. The banding and overall intensity vary across datasets and pipelines; Pipelines 2 and 6 are generally brighter than Pipelines 3 and 8.}
    \label{fig:app_ch5_posthoc_pipeline_grid}
\end{figure}

Pipeline choice changes the embedding geometry, but none of the shown variants closely resembles the DKT-derived graphs. Figure~\ref{fig:app_ch5_posthoc_pipeline_grid} provides the pipeline-by-dataset cosine-similarity comparison for \texttt{text-embedding-3-large}.

\end{document}